\documentclass{egpubl}
\usepackage{egsgp2021}

\usepackage[T1]{fontenc}
\usepackage{dfadobe}  

\usepackage{cite}  
\BibtexOrBiblatex
\electronicVersion
\PrintedOrElectronic
\ifpdf \usepackage[pdftex]{graphicx} \pdfcompresslevel=9
\else \usepackage[dvips]{graphicx} \fi

\usepackage{egweblnk} 

\usepackage[table,xcdraw,dvipsnames]{xcolor}
\usepackage{pgfplots} 
\usepackage{array}
\usepackage{multirow}
\usepackage{multicol}
\usepackage[utf8]{inputenc} 

\usepackage[acronym]{glossaries}
\usepackage{url}
\usepackage{algorithmic}
\usepackage{subcaption}
\usepackage{stmaryrd}
\usepackage{mathtools}
\usepackage{booktabs}
\usepackage[ampersand]{easylist}
\usepackage{rotating}
\usepackage{enumitem}
\usepackage{float}
\usepackage{comment}
\usepackage{balance}
\usepackage{placeins}
\usepackage{amsmath}
\usepackage{amsfonts}
\usepackage{amssymb}
\usepackage{tikz}
\usetikzlibrary{patterns,arrows.meta,positioning}
\usepackage{dsfont}
\usepackage{multicol}
\usepackage{mine}

\title{Scalable Surface Reconstruction\\ with Delaunay-Graph Neural Networks}

\author{\parbox{\textwidth}{\centering R. Sulzer$^{1,2}$\orcid{0000-0003-1707-7621}
L. Landrieu$^1$\orcid{0000-0002-7738-8141}
R. Marlet$^{2,3}$\orcid{0000-0003-1612-1758}
B. Vallet$^1$\orcid{0000-0002-9492-5180}
        }
        \\
{\parbox{\textwidth}{\centering $^1$ LASTIG, Univ Gustave Eiffel, ENSG IGN, F-94160 Saint-Mande, France\\
$^2$ LIGM, Ecole des Ponts, Univ Gustave Eiffel, CNRS, Marne-la-Vallée, France\\
$^3$ valeo.ai, Paris, France
      }
}
}
\begin{document}

\teaser{
\begin{center}
\vspace*{-2mm}
\begin{tabular}{@{}c@{}}
  \begin{tabular}{@{}ccc@{}}
    \setlength{\tabcolsep}{-15pt}
    \begin{subfigure}{.31\textwidth}
      \raisebox{-2mm}{\includegraphics[width=1\textwidth]{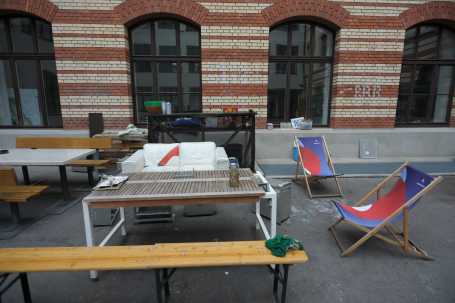}}
      \caption{Image from \emph{courtyard}.}
      \label{fig:court11}
    \end{subfigure}
    &
    \begin{subfigure}{.31\textwidth}
      \includegraphics[width=1\textwidth]{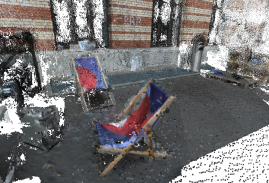}
      \caption{Associated MVS point cloud.}
        \label{fig:court12}
     \end{subfigure}
    &
    \begin{subfigure}{.31\textwidth}
      \includegraphics[width=1\textwidth]{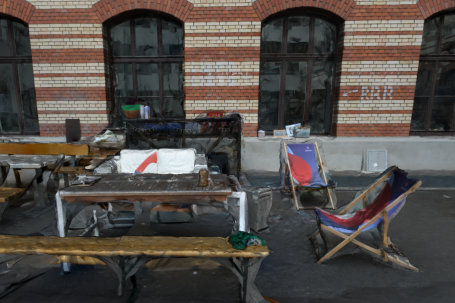}
      \caption{Our reconstruction textured.}
       \label{fig:court13}
     \end{subfigure}  
  \end{tabular}
  \\
  \begin{tabular}{@{}cccc@{}}
    \setlength{\tabcolsep}{-10pt}
    \begin{subfigure}{.23\textwidth}
      \raisebox{30mm}{}\includegraphics[width=1\textwidth]{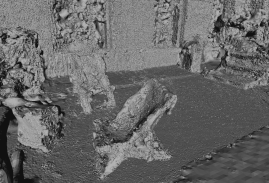}
      \caption{Screened Poisson.}
      \label{fig:court21}
    \end{subfigure}
    &
    \begin{subfigure}{.23\textwidth}
      \raisebox{30mm}{}\includegraphics[width=1\textwidth]{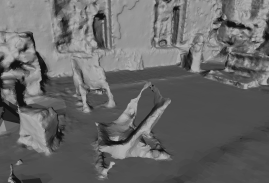}
      \caption{Vu \etal}
      \label{fig:court22}
     \end{subfigure}
    &
    \begin{subfigure}{.23\textwidth}
      \raisebox{30mm}{}\includegraphics[width=1\textwidth]{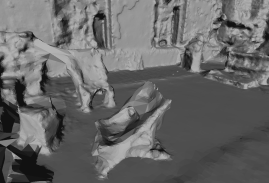}
      \caption{Jancosek \etal}
      \label{fig:court23}
     \end{subfigure}
     &
    \begin{subfigure}{.23\textwidth}
      \raisebox{30mm}{}\includegraphics[width=1\textwidth]{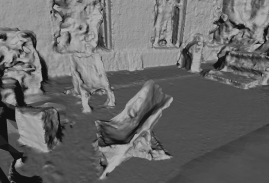}
      \caption{Ours.}
       \label{fig:court24}
     \end{subfigure}
  \end{tabular}
\end{tabular}
\vspace*{-2mm}
\caption{
{Reconstruction of the \emph{courtyard} scene of the ETH3D benchmark \cite{Schops2017}.
Top: a set of images, among which \Subref{fig:court11}, is transformed into a dense MVS point cloud pictured in \Subref{fig:court12}, from which our method reconstructs a mesh, displayed in \Subref{fig:court13} after texturation \cite{lettherebecolor}.
Bottom: we show untextured mesh reconstructions obtained by the screened Poisson algorithm in \Subref{fig:court21}, the algorithms of Vu \etal\cite{Vu2012} in \Subref{fig:court22} and Jancosek \etal\cite{Jancosek2011} in \Subref{fig:court23}, and our proposed reconstruction in \Subref{fig:court24}. Our method provides at the same time a higher accuracy (e.g., wall pattern in the background, that is reconstructed more truthfully) and a higher completeness (e.g., the back rest of the front chair).\vspace*{2mm}}
}
\label{fig:eth_courtyard}
\end{center}
}

\maketitle

\begin{abstract}
{
We introduce a novel learning-based, {visibility-aware,} surface reconstruction method for large-scale, defect-laden point clouds. Our approach can cope with the scale and variety of point cloud defects encountered in real-life Multi-View Stereo (MVS) acquisitions. Our method relies on a 3D Delaunay tetrahedralization whose cells are classified as inside or outside the surface by a graph neural network and {an energy model solvable with a} graph cut. Our model, making use of both local geometric attributes and line-of-sight visibility information, is able to learn a visibility model from a small amount of synthetic training data {and} generalizes to real-life acquisitions. 
{Combining the efficiency of deep learning methods and the scalability of energy-based {models}, 
our approach outperforms both learning and non learning-based reconstruction algorithms on two publicly available reconstruction benchmarks. Our code and data is available at \url{https://github.com/raphaelsulzer/dgnn}.}
}

\end{abstract}
%
\section{Introduction}\label{fig:teaser}
Reconstructing a surface from an unstructured point cloud is a long-standing and particularly challenging problem when applied to real-life acquisitions, due to occlusions, noise, outliers, non-uniform sampling, and misaligned scans \cite{Berger2016}.

A successful approach for dealing with large point clouds is to (i) tessellate the convex hull of the point cloud
using a \acrlong{3dt} (\acrshort{3dt}), (ii) label the resulting cells as \emph{inside} or \emph{outside}, and (iii) extract the surface as the interface between cells with different labels \cite{Jancosek2011,Labatut2009a,Vu2012}. This guarantees to produce non-self-intersecting and watertight surfaces, {a useful requirement for downstream engineering applications.
A remaining problem is that the methods typically rely on an energy formulation with  handcrafted unary and binary potentials. Tuning the balance between data fidelity and regularity in these methods tends to be difficult due to the high variability in nature and amplitude of the defects of real-life point clouds. 
}

\begin{figure*}[ht!]
\begin{tabular}{cccc}
\begin{subfigure}{0.23\textwidth}
\resizebox{\columnwidth}{!}{
\begin{tikzpicture}
[ray/.style={dotted,->, ultra thick, blue},
los/.style={ultra thick, red},
tri/.style={thick},
point/.style={circle, scale=.2, draw=black}
]
\node [point] at (0,0) (c0) {};
\node [point] at (2,0) (c1) {};
\node [point] at (4,0) (c2) {};
\node [point] at (1,2) (c3) {};
\node [point] at (3,-1) (c4) {};

\draw[tri] (c0) -- (c1) -- (c2) -- (c3) -- (c0) -- (c4) -- (c1) -- (c3);
\draw[tri] (c2) -- (c4);

\filldraw[color=black] (1.2,2.6) circle (0.15cm);
\fill[color=white] (1.2,2.6) circle (0.1cm);
\draw[very thick](0.9,2.6) .. controls (1,2.8) and (1.4,2.8) .. (1.5,2.6) .. controls (1.4,2.4) and (1,2.4) .. (0.9,2.6)--cycle;

\filldraw[color=black] (2.65,2.6) circle (0.15cm);
\fill[color=white] (2.65,2.6) circle (0.1cm);
\draw[very thick](2.35,2.6) .. controls (2.45,2.8) and (2.85,2.8) .. (2.95,2.6) .. controls (2.85,2.4) and (2.45,2.4) .. (2.35,2.6)--cycle;

\filldraw[color=black] (4.65,2.6) circle (0.15cm);
\fill[color=white] (4.65,2.6) circle (0.1cm);
\draw[very thick](4.35,2.6) .. controls (4.45,2.8) and (4.85,2.8) .. (4.95,2.6) .. controls (4.85,2.4) and (4.45,2.4) .. (4.35,2.6)--cycle;

\draw [los] (c1.center) -- +(+0.65cm,+2.6cm);
\draw [ray] (c1.center) -- +(-0.35cm,-1.4cm);

\draw [los] (c3.center) -- +(+0.2cm,+0.6cm);
\draw [ray] (c3.center) -- +(-0.85cm,-3.4cm);

\draw [los] (c2.center) -- +(+0.65cm,+2.6cm);
\draw [ray] (c2.center) -- +(-0.35cm,-1.4cm);

\draw [green, ultra thick, dashed] (0,2) to [out=00, in=180] (1,2) to [out=-30, in=120] (2,0) to[out=-30, in=180] (4,0);

\end{tikzpicture}
}
\caption{Delaunay triangulation and ray casting.}
\label{fig:visi}
\end{subfigure}
&
\begin{subfigure}{0.23\textwidth}
\resizebox{\columnwidth}{!}{
\begin{tikzpicture}
[ray/.style={dotted,->, ultra thick, blue},
los/.style={ultra thick, red},
tri/.style={thick, opacity = 0.3},
point/.style={circle, scale=.1, draw=none},
nod/.style={circle, scale=1, draw=black},
edgl/.style={[-{Straight Barb[left]}, ultra thick},
edgr/.style={[-{Straight Barb[right]}, ultra thick},
]
\node [point] at (0,0) (c0) {};
\node [point] at (2,0) (c1) {};
\node [point] at (4,0) (c2) {};
\node [point] at (1,2) (c3) {};
\node [point] at (3,-1) (c4) {};

\draw[tri] (c0) -- (c1) -- (c2) -- (c3) -- (c0) -- (c4) -- (c1) -- (c3);
\draw[tri] (c2) -- (c4);

\node [nod, fill = blue!80!red] at (1,0.66) (na) {};
\node [nod, fill = blue!90!red] at (1.66,-0.33) (nb) {};
\node [nod, fill = blue!20!red] at (2.33,0.66) (nc) {};
\node [nod, fill = blue!60!red] at (3,-0.33) (nd) {};

\draw [edgl] (na) to[out = 20, in = 160] (nc);
\draw [edgr] (nc) to[out = -160, in = -20] (na);

\draw [edgr] (nb) to[out = 20, in = 160] (nd);
\draw [edgl] (nd) to[out = -160, in = -20] (nb);

\draw [edgr] (na) to[out = -80, in = 150] (nb);
\draw [edgl] (nb) to[out = 100, in = -30] (na);

\draw [edgl] (nc) to[out = -80, in = 150] (nd);
\draw [edgr] (nd) to[out = 100, in = -30] (nc);

\end{tikzpicture}
}
\caption{Local and contextual learning with {gnn}.}
\label{fig:gnn}
\end{subfigure}
&
\begin{subfigure}{0.23\textwidth}
\resizebox{\columnwidth}{!}{
\begin{tikzpicture}
[ray/.style={dotted,->, ultra thick, blue},
los/.style={ultra thick, red},
tri/.style={thick, opacity = 0.3},
point/.style={circle, scale=.1, draw=none},
nod/.style={circle, scale=1, draw=black},
edg/.style={<->, ultra thick},
]
\node [point] at (0,0) (c0) {};
\node [point] at (2,0) (c1) {};
\node [point] at (4,0) (c2) {};
\node [point] at (1,2) (c3) {};
\node [point] at (3,-1) (c4) {};

\draw[tri] (c0) -- (c1) -- (c2) -- (c3) -- (c0) -- (c4) -- (c1) -- (c3);
\draw[tri] (c2) -- (c4);

\node [nod, fill = blue] at (1,0.66) (na) {};
\node [nod, fill = blue] at (1.66,-0.33) (nb) {};
\node [nod, fill = red] at (2.33,0.66) (nc) {};
\node [nod, fill = blue] at (3,-0.33) (nd) {};

\draw [edg] (na) -- (nb);
\draw [edg] (na) -- (nc);
\draw [edg] (nb) -- (nd);
\draw [edg] (nc) -- (nd);

\draw [green, ultra thick, dashed] (0,2) to [out=00, in=180] (1,2) to [out=-30, in=120] (2,0) to[out=-30, in=180] (4,0);
\end{tikzpicture}
}
\caption{Global optimization with graph cuts.}
\label{fig:graphcut}
\end{subfigure}
&
\begin{subfigure}{0.23\textwidth}
\resizebox{\columnwidth}{!}{
\begin{tikzpicture}
[ray/.style={dotted,->, ultra thick, blue},
los/.style={ultra thick, red},
tri/.style={thick, opacity = 0.3},
point/.style={circle, scale=.2, draw=black},
surface/.style={ultra thick, green}
]
\node [point] at (0,0) (c0) {};
\node [point] at (2,0) (c1) {};
\node [point] at (4,0) (c2) {};
\node [point] at (1,2) (c3) {};
\node [point] at (3,-1) (c4) {};

\fill [pattern=dots, pattern color=black] (c0.center) -- (c1.center) -- (c3.center);
\fill [pattern=dots, pattern color=black] (c0.center) -- (c1.center) -- (c4.center);
\fill [pattern=dots, pattern color=black] (c1.center) -- (c2.center) -- (c4.center);

\draw[tri] (c0) -- (c1) -- (c2) -- (c3) -- (c0) -- (c4) -- (c1) -- (c3);
\draw[tri] (c2) -- (c4);

\draw[surface] (c3) -- (c1) -- (c2);

\end{tikzpicture}
}
\caption{Surface reconstruction.\\~}
\label{fig:surf}
\end{subfigure}
\end{tabular}
\caption{\textbf{Pipeline: 2D representation of the different steps of our method.} \Subref{fig:visi} The input point cloud is triangulated, and visibility information is derived from lines of sight
\protect\begin{tikzpicture}
\protect\draw[very thick, red](0,0) -- (0.5,0);
\protect\draw[very thick, blue, ->, dotted](0.5,0) -- (1,0);
\protect\end{tikzpicture} 
and from camera positions
\protect\begin{tikzpicture}
\protect\filldraw[color=black] (0,0) circle (0.1cm);
\protect\fill[color=white] (0,0) circle (0.066cm);
\protect\draw[very thick](-0.2,0) .. controls (-0.13,0.13) and (0.13,0.13) .. (0.2,0) .. controls (0.13,-0.13) and (-0.13,-0.13) .. (-0.2,0)--cycle;
\protect\end{tikzpicture}.
\Subref{fig:gnn} A graph neural network uses this local and contextual visibility information to predict an occupancy score for each tetrahedron. \Subref{fig:graphcut} A global energy derived from the network's output finds a minimal cut 
\protect\tikz[baseline=-1mm]{
\protect\draw[color=green, dashed, very thick] (0,0.0) -- (0.5,0.0);} in an adapted flow graph. \Subref{fig:surf} The reconstructed surface  
\protect\tikz[baseline=-1mm]{
\protect\draw[color=green, very thick] (0,0.0) -- (0.5,0.0);}
is defined as the interface between cells with different (inside and outside) labels. 
}
\label{fig:steps}
\end{figure*}

In this paper, we present a novel method for reconstructing watertight surfaces from large point clouds based on a \acrshort{3dt} whose cells are associated with a graph-adjacency structure, local geometric attributes, and visibility information {derived from camera positions} (see Fig.\,\ref{fig:steps}).
We then train a \acrlong{gnn} (\acrshort{gnn}) to associate each cell with a probability of being inside or outside the reconstructed surface. 
{In order to obtain a spatially regular cell labelling, }
{these probabilities are incorporated into a global energy model that can be solved with a graph cut. This scheme directly predicts a spatially regular labeling, which leads to a smoother surface. Furthermore, graph-cut solving algorithms can easily scale to large point clouds, as opposed to other learning-based surface reconstruction methods, which tend to be limited to objects, or operate with sliding windows, as remarked by \cite{Peng2020}. }


{To the best of our knowledge, our method is the first deep-learning-based mesh reconstruction algorithm able to take visibility information into account. This property is valuable, especially in areas lacking sufficiently dense input points. 
It is also the first deep learning surface reconstruction method using a memory-efficient \acrshort{gnn} implementation built on a \acrshort{3dt}. We argue that combining the scalability of traditional computational geometry algorithms with the adaptability of modern deep learning approaches paves the way to learning-based large-scale 3D information processing.}
{We validate our approach by showing that, even when trained on a small synthetic dataset, our method is able to generalize to large-scale, real-life, and complex 3D scenes and reach state-of-the-art performance on an open-access MVS dataset \cite{Schops2017} (see Fig.\,\ref{fig:teaser}).}
\section{Related Work}
\subsection{Graph-Cut-Based Surface Reconstruction}
Visibility-based surface reconstruction from LiDAR scans and/or \acrshort{mvs} data is traditionally formulated as a graph-cut optimization problem \cite{Bodis-Szomoru2017,Caraffa2017,Jancosek2011,Jancosek2014,Vu2012,Zhou2019}.
The 3D space is discretized into the cells 
of a \acrshort{3dt} of captured points \cite{Labatut2009a} or the cells of an arrangement of detected planes \cite{ChauveCVPR2010}. {A graph $(\cT,\cE)$ is formed {for which} the vertices are the cells $\cT$ of the complex, and the edges in $\cE$ connect cells with a common facet.} 
{Each cell $t \in \cT$ is to be assigned with an occupancy label $l_t$ in $\{0,1\}$, where $0$ means outside and $1$ means inside. For this,}
{each 
{cell} 
$t$ is attributed a \emph{unary potential} $U_t$ expressing a likelihood of being inside or outside the scanned object. {Additionally,} each facet interfacing two {adjacent cells} 
$s$ and $t$ is attributed a \emph{binary potential} $B_{s,t}$,
{which takes low values when the facet is likely to be part of {a regular} reconstructed surface {and higher values otherwise}.}
{The label assignment of cells is performed}
by minimizing an energy in the following form:}
\begin{align}
    E(l) = \sum_{t \in \cT} U_t (l_t) + \lambda \sum_{(s,t) \in \cE} B_{s,t}(l_s, l_t)~,
    \label{eq:ising}
\end{align}
where $\lambda \geq 0$ is the regularization strength. 
This energy $E$ is globally minimized by computing a minimum cut in an appropriate flow graph or using a linear programming approach \cite{BoulchCGF2014}.

The unary potentials commonly depend on visibility criteria, such as: (i)~{cells with sensors} 
are always outside, (ii)~{cells traversed by lines of sight (virtual lines between a sensor and an observed point) are likely outside, or (iii)}~cells \emph{behind} a point are likely inside.
These visibility models are not robust to the acquisition noise and outliers of {real-life} point clouds, so the unary potentials can be adjusted to the local point density \cite{Vu2012,Jancosek2011,Jancosek2014,Zhou2019}, or by using other modalities \cite{Bodis-Szomoru2017}.

Binary potentials are used to force neighbouring cells crossed by the same line of sight 
to have the same labeling. Additionally, they can incorporate low-area \cite{Bodis-Szomoru2017,Caraffa2017} or other shape-based priors \cite{Jancosek2011, Labatut2009a, Vu2012}.
Instead of hand-tuning the visibility model, we propose to learn it by training a neural network to produce unary potentials from 
{local visibility and {local} geometric information.}

\subsection{Deep Learning-Based Surface Reconstruction}
{Recently, deep learning-based models have been proposed for reconstructing surfaces from point clouds or other modalities, operating on a discrete mesh or with continuous 
functions.}

{\noindent \textbf{Surface-based approaches} rely on {transforming} a discretized 2D surface, such as 2D patches or spheres \cite{Groueix2018, yang2018foldingnet, pointtrinet, liu2020}, meshes \cite{Gkioxari_2019, point2mesh, dai2019scan2mesh}, charts \cite{Williams2018}, or learned primitives \cite{deprelle2019learning}, in order to best fit an input point cloud.
While such methods can lead to impressive visual results, they either cannot guarantee that the output mesh is watertight and intersection-free, or are limited to simple topologies and low resolution. Additionally, they are typically memory intensive, which prevents them from scaling to large scenes.
}


\noindent {\textbf{Volumetric approaches} 
learn a continuous mapping from the input space $\bR^3$ either to $\bR$, defining the signed distance to the surface \cite{Gropp2020, Park2019, Atzmon_2020_CVPR, Atzmon_2021_ICLR, deepLS}, or directly to an occupancy value $\{0, 1\}$ \cite{Mescheder2019, Mi2020, Peng2020}. 
The network training can be either unsupervised \cite{liu2019learning}, aided by geometric regularization \cite{Gropp2020}, or supervised by ground-truth surface information \cite{Mescheder2019,Mi2020, Peng2020}.}
{
Some continuous methods \cite{Park2019, Mescheder2019, Peng2020} predict the occupancy or signed distance conditionally to a latent shape representation, and thus learn a dataset-specific shape distribution. This can lead to difficulty in generalizing to shapes from unseen classes.}

{Even though volumetric approaches define a surface in continuous space with implicit functions, they often rely on a discretization of 3D space to learn these functions \cite{Mescheder2019, Mi2020, Peng2020}.}
{Recent works propose to scale these methods to larger scenes using an octree structure \cite{Mi2020} or a sliding window strategy \cite{Peng2020}.}



{While our method also relies on a discretization of space, our \acrshort{3dt} is directly computed from the input point cloud and is thus adaptive to the local resolution. Our method guarantees to produce watertight surfaces, can operate at large scale, and generalizes to unseen shapes and scenes.}

\section{Methodology}

We explain here how to construct a \acrshort{3dt} augmented with expressive but lightweight visibility features that are leveraged by a memory-efficient \acrshort{gnn} and used in a global energy formulation to extract the target surface.
\subsection{Visibility-Augmented 3D Tetrahedralization}
We consider $\cP \in \bR^{3\times P}$ a 3D point cloud defined by the absolute point positions in space, and $\cC \in \bR^{3\times C}$ the absolute positions of a set of cameras used to capture these points. 
\RAPH{We first construct a \acrshort{3dt} tessellating the convex hull of $\cP$ into a finite set of tetrahedra $\cT$.}
Each tetrahedron $t$ is characterized by its four vertices $\cV_t \in \bR^{3 \times 4}$ \RAPH{and four facets $\cF_t \in \bN^{3 \times 4}$. At the boundary of the convex hull, each facet is incident to an infinite cell whose fourth vertex is at infinity.
This ensures that each facet of the \acrshort{3dt} is incident to exactly two tetrahedra.}

Let $\cL \subset \cC\,{\times}\,\cP$ be the \emph{lines-of-sight} from cameras $c$ of $\cC$ to points $p$ of $\cP$ seen from~$c$.
{A \emph{line-of-sight} $\los{c}{p} \in \cL$ is an oriented segment from $c$ to $p$.}
In the case of \acrshort{mvs} point clouds, a single point can be seen from  multiple cameras.
%
{Similarly, we call $\ray{c}{p} \in \cR \subset \cC\,{\times}\,\cP$ the \emph{ray} extending line-of-sight $\los{c}{p}$ from the seen point $p$ to infinity.}
{To simplify the computation of visibility information, we truncate the ray traversal after the second tetrahedron. {For instance, in \figref{fig:ray_features}, $\ray{c}{p}$ does not go beyond $t_3$.}} 

\subsection{Feature Extraction}
%
\label{sec:handcrafted}
{The \emph{occupancy}, or insideness, of a tetrahedron \textit{w.r.t.} the target surface can be inferred by combining geometrical and visibility information. Indeed, a tetrahedron $t$ traversed by a line-of-sight $\los{c}{p}$ is \emph{see-through}, and most likely lies outside the surface. Conversely, if a tetrahedron is traversed by a ray $\ray{c}{p}$ and no line-of-sight, it may lie inside the surface, 
{especially if}
 close to $p$.}

However, visibility-based information 
is not sufficient to retrieve a perfect labelling of tetrahedra.
First, there is no connection between the discretization of the space by the \acrshort{3dt} and the distribution of lines-of-sight. There may be a significant number of tetrahedra not traversed by any line-of-sight nor any ray, depending on the geometry of the acquisition. 
Second, noise and outliers --- stemming from \acrshort{mvs}
{for example} ---
can result in inaccurate and unreliable visibility information.
{Thus, we propose to use a \acrshort{gnn} to propagate and smooth visibility-based {information, as well as} other contextual information, to all tetrahedra in the \acrshort{3dt} of an object or a scene.}
%
%

\noindent\textbf{Tetrahedron features}
While one could argue for directly learning features from tetrahedra and camera positions in an end-to-end fashion, this resulted in our experiments in a significant computational overhead and a very difficult geometric task for a neural network to learn. Instead, we propose to derive computationally light, yet expressive, handcrafted features encoding the local geometry and visibility information of tetrahedra.

For a tetrahedron $t \in \cT$, we denote by $\cLv_t$ the set of lines-of-sight that traverse $t$ and that end at one of its vertices $v \in \cV_t$, and by $\cLf_t$ the set of lines-of-sight that intersect $t$ through its facets and do not end at one of its vertices:
\begin{align}
    \cLv_t & = \{(c,p) \!\in\! \cL \mid (\los{c}{p}) \cap t \neq \varnothing,~ p \in \cV_t \} \\
    \cLf_t & = \{(c,p) \!\in\! \cL \mid (\los{c}{p}) \cap t \neq \varnothing,~ p \notin \cV_t \}\rlap.
\end{align}
Likewise, we denote $\cRv_t$ and $\cRf_t$ the equivalent sets for rays in $\cR$. These definitions are illustrated on \figref{fig:ray_features}.
\begin{figure}[t]
\centering
\begin{tabular}{c}
\resizebox{1\columnwidth}{!}{
\begin{tikzpicture}
[ray/.style={dotted,->, ultra thick},
los/.style={ultra thick}]
\node [circle, scale = 0.5, fill = black!10!white] at (0,0) (c0) {};
\node [circle, scale = 0.3, fill = black!10!white] at (-2,-2) (c1) {};
\node [circle, scale = 0.3, fill = black!10!white] at (-2,2) (c2) {};
\node [circle, scale = 0.3, fill = black!10!white] at (-4,-2) (c3) {};
\node [circle, scale = 0.3, fill = black!10!white] at (2,-2) (c4) {};
\node [circle, scale = 0.3, fill = black!10!white] at (2,+2) (c5) {};
\node [circle, scale = 0.3, fill = black!10!white] at (4,2) (c6) {};

\fill [fill = orange, opacity = 0.2] (c0.center) -- (c1.center) -- (c2.center) -- (c0.center);
\fill [fill = cyan, opacity = 0.2] (c1.center) -- (c2.center) -- (c3.center) -- (c1.center);
\fill [fill = black, opacity = 0.2] (c2.center) -- (c0.center) -- (c5.center) -- (c2.center);
\fill [fill = purple, opacity = 0.2] (c0.center) -- (c4.center) -- (c5.center) --  (c0.center);
\fill [fill = black, opacity = 0.2] (c1.center) -- (c0.center) -- (c4.center) -- (c1.center);
\fill [fill = ForestGreen, opacity = 0.4] (c4.center) -- (c5.center) -- (c6.center) --  (c4.center);

\fill [pattern=dots, pattern color=black!50!white] (c2.center) to[out = -30, in = 90] (c0.center) to[out = -90, in = 30] (c1.center) to (c4.center) to (c6.center) to (c5.center) to (c2.center);
\draw [ultra thick, ForestGreen] (c2.center) to[out = -30, in = 90] (c0.center) to[out = -90, in = 30] (c1.center);

\draw [very thick, black!50!white] (c3) -- (c2) -- (c1) -- (c0) -- (c2) -- (c5) -- (c6) -- (c4) -- (c5) -- (c0) -- (c4);
\draw [very thick, black!50!white] (c4) -- (c1) -- (c3);

\node [draw = none] at (+2.7,0.8)  {\Large $t_3$};
\node [draw = none] at (+1.5,0.8)  {\Large $t_2$};
\node [draw = none] at (-1.5,-0.8)  {\Large $t_1$};
\node [draw = none] at (-2.7,-0.8)  {\Large $t_0$};
\node [draw = none] at (0.3,1.4)  {\Large $t_4$};
\node [draw = none] at (0.3,-1.4)  {\Large $t_5$};

\node [draw = none, right] at (0,-0.4)  {\Large $p$};

\filldraw[color=black] (-5,0) circle (0.3cm);
\fill[color=white] (-5,0) circle (0.2cm);
\draw[very thick](-5.6,0) .. controls (-5.4,0.4) and (-4.6,0.4) .. (-4.4,0) .. controls (-4.6,-0.4) and (-5.4,-0.4) .. (-5.6,0)--cycle;

\node [draw = none] at (-5,-0.6)  {\Large $c$};

\coordinate[ultra thick, left = 15cm of c0] (eye);
\coordinate (losinter12) at (intersection of c1--c2 and c0--eye);
\coordinate (losinter23) at (intersection of c2--c3 and c0--eye);

\coordinate[above = 0.3cm of losinter12] (losinter12up);
\coordinate[above = 0.3cm of c0.center] (c0up);

\coordinate[above = -0.6cm of losinter23] (losinter23down);
\coordinate[above = -0.6cm of c0.center] (c0down);

\draw[-, dashed, thick, red]  (losinter23down) -- (losinter23);
\draw[-, dashed, thick, red]  (losinter12up) -- (losinter12);
\draw[-, dashed, thick, red]  (c0up) -- (c0);
\draw[-, dashed, thick, red]  (c0down) -- (c0);

\draw[{Latex[length=0.01mm, width=3mm]}-{Latex[length=0.01mm, width=3mm]}, thick, red]  (losinter12up) -- (c0up) node [above left = 0mm and 8mm,  fill=none, text = red] {$\small{\text{len}(\los{c}{p},t_1)}$};
\draw[{Latex[length=0.01mm, width=3mm]}-{Latex[length=0.01mm, width=3mm]}, thick, red]  (losinter23down) -- (c0down) node [above left = 0mm and 8mm,  fill=none, text = red] { $~\;\small{\text{len}(\los{c}{p},t_0)}$};

\coordinate[right = 5cm of c0] (x);
\coordinate (rayinter) at (intersection of c4--c6 and c0--x);

\draw [ray, draw=blue] (c0.center) -- (rayinter) node [below right,  fill=none, text = blue] {\Large $\ray{c}{p}$};
\draw [los, draw=red ] (c0.center) -- +(-5cm,0cm) node [above, near end,  fill=none, text = red] {\Large $\los{c} {p}$};

\end{tikzpicture}
} \\
\begin{tabular}{llll}
   $\textcolor{red}{\los{c}{p}} \in \textcolor{cyan}{\cLf_{t_0}}$  & $\textcolor{red}{\los{c}{p}} \in \textcolor{orange}{\cLv_{t_1}}$ &
   \multirow{2}{*}{
    \resizebox{!}{0.04\textheight}{
   \begin{tikzpicture}
   \fill [pattern=dots, pattern color=black!50!white] (0,1) to[out = -30, in = 90] (+0.3,0) to[out = -90, in = 30] (0,-1) to (1,-1) to (1,1) to (0,1);
\draw [ultra thick, ForestGreen] (0,1) to[out = -30, in = 90] (+0.3,0) to[out = -90, in = 30] (0,-1);
\end{tikzpicture}
}
}
&
real\\
    $\textcolor{blue}{\ray{c}{p}} \in \textcolor{purple}{\cRv_{t_2}}$  & $\textcolor{blue}{\ray{c}{p}} \in \textcolor{ForestGreen}{\cRf_{t_3}}$ &
    &
    surface
\end{tabular}
\end{tabular}
\caption{\textbf{Visibility Model.} 2D representation of our visibility features. A line-of-sight $\textcolor{red}{\los{c}{p}}$ between a camera $c$ and a visible 3D point $p$ also defines a ray $\textcolor{blue}{\ray{c}{p}}$. 
The line-of-sight $\textcolor{red}{\los{c}{p}}$ traverses the two outside 
tetrahedra $\textcolor{cyan}{t_0}$ and $\textcolor{orange}{t_1}$, while the ray $\textcolor{blue}{\ray{c}{p}}$ traverses the two inside tetrahedra $\textcolor{purple}{t_2}$ and $\textcolor{ForestGreen}{t_3}$. Neither $\textcolor{red}{\los{c}{p}}$, nor $\textcolor{blue}{\ray{c}{p}}$ traverse $\textcolor{gray}{t_4}$ or $\textcolor{gray}{t_5}$; they thus do not contribute to their visibility information.}
\label{fig:ray_features}
\end{figure}
These sets are informative for determining the {occupancy of a tetrahedron.}
Indeed, a tetrahedron $t$ for which  $\cRv_t$ is nonempty indicates that it is directly \emph{behind} an element of the surface, hinting at a higher probability of insideness. A nonempty $\cRf_t$ indicates that $t$ was hidden by a surface, hinting at a possible insideness. Indeed, since the \emph{hit} occurred \emph{before} $t$, this carries less confidence as it could be due to an occlusion or a thin structure. 

Conversely, a tetrahedron $t$ with nonempty $\cLf_t$ indicates that it is traversed by a line-of-sight, indicating a high probability of outsideness. A nonempty $\cLv_t$ also indicates that $t$ is traversed by a line-of-sight, but since the \emph{hit} is on one of the corners of the tetrahedron, this prediction is likely to be affected by acquisition noise, and hence {has a lower confidence}. 

{To characterize the influence of lines-of-sights and rays with respect to a given tetrahedron $t$, we define two measures: $\text{count}(t)$ and $\text{dist}(t)$.
$\text{count}(t) \in \mathbb{N}^4$ corresponds to the number of each type of lines or rays intersecting with $t$:
\begin{align}
\text{count}(t)= \left[\card{\cLv_t}, \card{\cLf_t}, \card{\cRv_t}, \card{\cRf_t} \right]~.
\end{align}
Then, to measure the \emph{proximity} between $t$ and the impact point $p$ of a traversing line-of-sight $\los{c}{p}$
, we define $\text{len}(\los{c}{p},t)$ as the distance between $p$ and the exit point of $\los{c}{p}$ in $t$ seen as from $p$. {As represented in \figref{fig:ray_features}}, this corresponds to the length of the longest segment between $p$ and the portion of $\los{c}{p}$ intersecting $t$:
\begin{align}
\label{eq:labatut_d}
\text{len}(\los{c}{p}, t) = \max_{y \in (\los{c}{p}) \cap t} \lVert p - y\rVert~.
\end{align}
\LOIC{When $(\los{c}{p}) \cap t$ is empty, $\text{len}(\los{c}{p}, t)$ is set to zero.} We define $\text{len}(\ray{c}{p},t)$ in the same manner for rays.
{Finally,} $\text{dist}(t)$ characterizes the proximity of tetrahedron $t$ with the observed points $p$ of its intersecting lines-of-sight and rays:
\begin{align}\nonumber
\text{dist}(t)= &\left[
\min_{\los{c}{p}\in \cLv_t}\text{len}(\los{c}{p},t),
\min_{\los{c}{p}\in \cLf_t}\text{len}(\los{c}{p},t),\right.\\
&\left.\min_{\ray{c}{p}\in \cRv_t}\text{len}(\ray{c}{p},t),
\min_{\ray{c}{p}\in \cRv_t}\text{len}(\ray{c}{p},t)
\right]~.
\end{align}
We complement the $8$ visibility features defined by $\text{count}(t)$ and $\text{dist}(t)$ with 4 morphological features: the volume of $t$, the length of its shortest and longest edges, and the radius of its circumsphere. This leads to a set of $12$ handcrafted features $f_t$ for each tetrahedron $t \in \cT$, that we normalize (zero mean and unit standard deviation) independently.
}

It is important to note that none of the aforementioned features can be computed in a meaningful way for infinite cells of the \acrshort{3dt}. We simply set all feature values to zero, which can be interpreted as a padding strategy.
\subsection{Contextual Learning}
\label{sec:contextual}
\begin{figure}[t!]
\centering
\begin{tabular}{c}
   \includegraphics[width=.8\columnwidth]{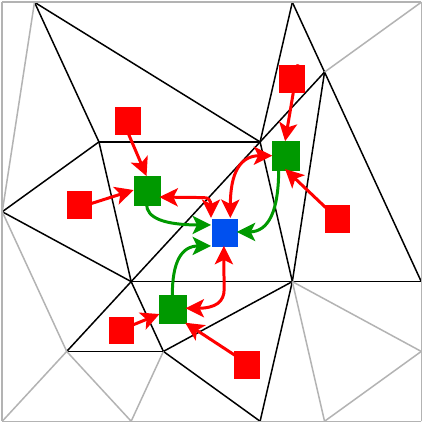}\\
  \begin{tikzpicture}
    \node[rectangle,fill=red,scale=2] at (0,.5) (n2) {};
    \node[rectangle,fill=green!60!black,scale=2] at (2,.5) (n3) {};
    \node[rectangle,fill=blue,scale=2] at (4,.5) (n4) {};
    \node[rectangle,fill=none,draw=none,scale=1] at (6,.5) (n5) {$i/o$};
    
    \node[rectangle,fill=none] at (0,0) (n11) {$12$};
    \node[rectangle,fill=none] at (2,0) (n21) {$128$};
    \node[rectangle,fill=none] at (4,0) (n22) {$128$};
    \node[rectangle,fill=none, draw=none] at (6,0) (n23) {$2$};
    \draw[ultra thick,-{Latex[length=3mm, width=4mm]}, red] (n2)--(n3);
    \draw[ultra thick,-{Latex[length=3mm, width=4mm]}, green!60!black] (n3)--(n4);
    \draw[ultra thick,-{Latex[length=3mm, width=4mm]}, black] (n4)--(n5);
  \end{tikzpicture} 
\end{tabular}
\caption{
{\textbf{Graph Convolution Scheme.} Illustration of our \acrshort{gnn} update for tetrahedron 
\protect\begin{tikzpicture}
 \protect\node[rectangle,fill=blue,scale=1] at (0,0) (n4) {}; 
\end{tikzpicture}. 
Information is pooled at different hops (here $K=2$) into an increasingly rich descriptor. A linear layer based on the last cell embedding assigns an inside/outside score to the central tetrahedron. {Note that each prediction can be performed independently for each tetrahedron by considering only the $K$-hop subgraph.}
}
}
\label{fig:gcn}
\end{figure}

We learn contextual information with a \acrshort{gnn} \RAPH{using} 
the propagation scheme GraphSAGE of Hamilton \etal\cite{Hamilton2017} with a depth of $K$ (see \figref{fig:gcn}). \LOIC{This scheme can be performed independently for each tetrahedron, allowing us to perform inference on large graphs \LOIC{with limited memory requirements.}}

We denote by $G=(\cT,\cE)$ the undirected graph whose edges $\cE \subset \cT^2$ link cells that are adjacent, i.e., share a facet. 
We consider one tetrahedron $t$ in $\cT$, and compute $\text{hop}(t,K)$ its $K$-hop neighborhood in $G$, i.e., 
\LOIC{the set of nodes $s$ of $\cT$ which can be linked to $t$ using at most $K$ edges. }
%
\LOIC{We leverage the local context of a tetrahedron $t$ with a message-passing scheme over its local neighborhood in $G$.
We first initialize the features of all nodes $s$ in the subgraph $\text{hop}(t,K)$ with the handcrafted features defined in \secref{sec:handcrafted}: $x^{0}_s = f_s$. We then apply the following update rule in two nested loops over $k=0,\ldots,K-1$ and for all $s \in \text{hop}(t,K-1)$:
\begin{align}
    x^{k+1}_s & = \sigma\left(\text{norm}\left(W^{(k)} \left[x^{k}_s \;\middle|\middle|\; \mean_{u \in \mathcal{N}(s)}\left(x^{k}_u\right) \right]\right)\right)~,
\end{align}
with $\mathcal{N}(s)$ the one-hop neighborhood of node $s$, $\sigma$ an activation layer, norm a normalization layer, and $[\cdot \mid\mid \cdot]$ the concatenation operator.
$\{W^{(k)}\}_{k=0}^{K-1}$ is a set of $K$ learned matrices, each operating only at the $k$-th iteration. After $K$ iterations, a multilayer perceptron (MLP) maps the embedding $x^K_t$ to a vector of dimension $2$ representing the inside/outside scores for tetrahedron $t$:
\begin{align}
    (\ip_t, \op_t) = \text{MLP}(x^K_t)~.
\end{align}
}

The main advantage of this simple scheme is that it can be performed \emph{node-wise} from the $K$-hop neighborhood of each node and run the update scheme locally. Memory requirements only depend on $K$, i.e., subgraph extraction, {and not on the size of the full graph $G$}.
This allows us to scale inference to large graphs.
{Likewise, training can be done by sampling subgraphs of depth at least $K$, and does not require to load large graphs in memory.}

\subsection{Loss Function}
For defect-laden point clouds affected by noise, the ground-truth surface is generally not exactly aligned with the faces of the 3DT created from the input points.
Consequently, tetrahedra intersecting the true 
surface can be only partially inside or outside, and cannot be attributed a \emph{pure} $0$ or $1$ occupancy label. Instead, we define the ground-truth insideness/outsideness $\ig \in [0,1]^{\card{\cT}}$ as the proportion of each cell's volume lying inside of a ground-truth {closed object}; $\ig$ can take any value between $0$ and~$1$.

\LOIC{We convert the tetrahedra's predicted inside/outside scores to an occupancy using the sotfmax fonction: }
\LOIC{
\begin{align}
   \is_t = \frac{\exp(\ip_t)}{\exp(\ip_t)+\exp(\op_t)}
\end{align}
}
\LOIC{We define the fidelity loss for each $t$ as the Kullback-Leibler divergence of the true occupancy $\ig_t$ and the softmaxed predicted occupancy $\is_t$:
\begin{align}
    \LKL_t\left(\is_t\right) & = \ig_t \log\left(\is_t\right) + (1-\ig_t) \log\left(1-\is_t\right)+q~,
\end{align}
with $q$ a {quantity} 
that does not depend on $\is_t$, and can thus be ignored while training the network.
We define the total loss as the average of all tetrahedra's fidelity weighted by 
their volume $V_t$:
\begin{align}
    L\left(\is\right) & = \frac1{\sum_{t \in \cT}V_t}\sum_{t \in \cT}V_t \:\LKL_t\left(\is_t\right)~.
\end{align}
} 


\if 1 0
\subsection{Spatial Regularization}
In order to implement the prior that the reconstructed surface should be simple and with a low area, we define a spatially regularizing penalty $\LTV$ as the weighted total variation of the insideness:
\begin{align}\nonumber
    \LTV\left(\ip\right) & = \sum_{(s,t) \in E} \beta_{s,t} \lVert \ip_s - \ip_t \rVert~,
\end{align}
with $\beta \in \cR_+^\vert E\vert$ defined as:

cc-weight (=0.3) + area-weight * surface area (normalized by mean area of the scene) + angle-weight (=0.1) * $\beta$-skeleton

\begin{align}\nonumber
\beta_{s,t} = \gamma_0 + \gamma_1 S(s,t)
\end{align}

The total loss used to train the neural network is defined as:
\begin{align}\nonumber
L(\ip) = \LKL(\ip) + \lambda \LTV(\ip)~,
\end{align}
with $\lambda>0$ the regularization strength, defined in the rest of the paper as $0.1$. 
\fi
\subsection{Global Formulation }
\label{sec:global}
Defining the target surface directly from the inside/outside scores predicted by the network can result in a jagged surface due to non-consistent labelling of neighboring tetrahedra. To achieve a smooth label assignment, even in areas with heavy noise, we use the inside/outside scores $i_t,o_t$ to define the unary potentials in the formulation of \equaref{eq:ising}:
%
%

\LOIC{
with $[x=y]$ the Iverson bracket, equal to $1$ if $x=y$ and $0$ otherwise.}
%
%

%

{Following the idea of Labatut \etal \cite{Labatut2009a}, we define the binary potentials introduced in \equaref{eq:ising} with a surface quality term that allows us to reconstruct a smooth surface and to efficiently remove isolated or non-manifold components in the final surface mesh. See the supplementary for more details. We also add a constant factor $\alpha_\text{vis}$ to $o_t$ for tetrahedra containing a camera, indicating that they must lie outside the surface.}
{The energy $E(l)$ in Eq.\,\eqref{eq:ising} with unary and binary potentials as defined above can be minimized efficiently by constructing a flow graph and using a min-cut solver \cite{boykov2004experimental}.}
%
\subsection{Surface Extraction and Cleaning}
\label{sec:clean}


{We can define the target  surface by considering the labeling of $\cT$ obtained by minimizing $E(l)$. The reconstructed surface is composed of all triangles whose adjacent tetrahedra have different labels. Triangles are oriented towards the outside tetrahedra. }
{For open scene reconstruction, we optionally apply a standard mesh cleaning procedure, implemented in OpenMVS \cite{cernea2015openmvs}, by removing spurious and spike faces (whose edges are too long). This is especially useful {for outdoor scenes containing areas with} very little input data, such as far-away background or sky, and for which 
outliers can result in isolated components with low-quality surfaces.}
{In our experiments,} all competing methods, at least visually, benefit from this classic postprocessing for open scene reconstruction. 
\section{Experiments}
{In this section, we present the results of two {sets of} numerical experiments to show the performance of our reconstruction method for both objects and large-scale scenes. In both settings, our method is only trained on a small synthetic dataset, and yet outperforms state-of-the-art learning and non learning-based methods, highlighting its high capacity for generalization.}
\subsection{Evaluation Setting}
\label{sec:evaluation}

\noindent\textbf{Training Set.}
We train our network on a small subset of 10 shapes for each of the 13 classes of the ShapeNet subset from \cite{choy20163dr2n2}.
\LOIC{We found this small number to be sufficient for our network to learn diverse local shape configurations.}
We produce watertight meshes of these models using the method of Huang \etal~\cite{huang2018shapenetwatertight}. We then synthetically scan the models {with different degrees of outliers and noise} as described in the supplementary material, and build corresponding 3DTs. {To obtain the ground truth occupancy, we randomly sample} $100$ points in each tetrahedron, and determine the percentage of these sampled points lying inside the ground-truth model.

\noindent\textbf{Hyper-Parameters.}
We train our model by extracting batches of $128$ subgraphs of depth $K=4$ centered around random tetrahedra of our training set. We parameterize our model with $K=4$ linear layers of width $64$, $128$, $256$ and $256$ for each hop respectively. \RAPH{After each linear layer, we apply batch normalization \cite{batchnorm} and ReLU non-linearities. The final cell embeddings are mapped to inside/outside scores using an MLP 
$256\rightarrow 64 \rightarrow 2$.}
We train the network with the Adam optimizer \cite{kingma2017adam} with an initial learning rate of $10^{-4}$ which we decrease by a factor of $10$ every $10$ epochs. 
For the graph-cut optimization, we set the camera bias term $\alpha_{vis}$ to $100$ and the regularization strength to $\lambda = 1$.

{We use the same hyper-parameters for all {dataset variants, in particular for all} settings of the scanning procedure of Berger et al.'s benchmark \cite{Berger2011}. While we could choose parameters that better fit specific noise and outlier levels, we argue that a single real-life scene can present multiple defect configurations simultaneously. Consequently, reconstruction algorithms should be versatile enough to handle different noise and outlier ratios with a single parameterization.}

\begin{table*}[t!]
\caption{\textbf{Quantitative results on Berger et al.'s dataset.} Object-level reconstruction with various point cloud settings: low resolution (LR), high resolution (HR), high resolution with added noise (HRN), high resolution with added outliers (HRO), high resolution with noise and outliers (HRNO). We measure the symmetric Chamfer distance to the ground truth (per-point average for objects of size 75 as done in Berger et al.'s benchmark \cite{Berger2011}), volumetric IoU~(\%), number of components (ground-truth meshes all have only one component) and number of non-manifold edges (none in the ground truth). All metrics are averaged over the 5 shapes of the benchmark dataset. We compare IGR \cite{Gropp2020} ``trained'' on each of the 5 variants (LR, HR, HRN, HRO, HRNO) of the 5 shapes, screened Poisson reconstruction \cite{screened_poisson} with an octree of depth 10, Labatut \etal\cite{Labatut2009a} with $\sigma$ set according to an estimation of the scan noise and ConvONet \cite{Peng2020} and our method trained on the ShapeNet subset from \cite{choy20163dr2n2}.}
\centering
\begin{tabular}{l@{~}r|cccccc|cccccc}
\toprule
& & \multicolumn{6}{c|}{\textbf{Chamfer distance} (per-point ave. \%) ~~[$\downarrow$]}
& \multicolumn{6}{c}{\textbf{Volumetric IoU} (\%)~~[$\uparrow$]}
\\
\bf Method & & LR & HR & HRN & HRO & HRNO & Mean
& LR & HR & HRN & HRO & HRNO & Mean \\
\midrule
\bf ConvONet & \cite{Peng2020}          &           1.90&	        1.80&	        2.31&	        2.91&	        3.73&	        2.53&	        67.8&	        71.3&	        62.9&	        61.4&	        57.3&	        64.1\\
\bf IGR & \cite{Gropp2020}              &           1.03&	        0.44&	        0.80&	        11.87&	        11.50&	        5.13&	        80.4&	        93.0&	        84.2&	        27.5&	        27.8&	        62.6      \\
\bf Poisson & \cite{screened_poisson}   &           1.09&	        0.48&	        0.80&	        0.46&	        0.86&	        0.74&	        79.1&	        91.9&	        84.3&	        91.9&	        83.3&	        86.1 \\
\bf Labatut et al. & \cite{Labatut2009a}&           0.89&	        0.42&	        0.89&	        0.46&	        0.95&	        0.72&	        81.9&	        94.5&	        80.9&	        94.3&	        80.6&	        86.4 \\
\bf Ours &                              &  \textbf{0.88}&	\textbf{0.41}&	\textbf{0.77}&	\textbf{0.41}&	\textbf{0.78}&	\textbf{0.65}&	\textbf{82.0}&	\textbf{95.6}&	\textbf{84.7}&	\textbf{95.3}&	\textbf{84.7}&	\textbf{88.5} \\
\midrule
& & \multicolumn{6}{c|}{\textbf{Number of components} ~~[$\downarrow$]}
& \multicolumn{6}{c}{\textbf{Number of non-manifold edges}~~[$\downarrow$]}
\\
\bf Method & & LR & HR & HRN & HRO & HRNO & Mean
& LR & HR & HRN & HRO & HRNO & Mean \\
\midrule
\bf ConvONet &\cite{Peng2020}          &            3.2&	        2.0&	        6.0&	        12.8&	        14.0&	        7.6&	\textbf{0.0}&    \textbf{0.0}&	\textbf{0.0}&	\textbf{0.0}&	\textbf{0.0}&	\textbf{0.0} \\
\bf IGR & \cite{Gropp2020}             &            2.2&	        2.2&	        43.0&	        43.0&	       101.2&	        38.3&	\textbf{0.0}&	\textbf{0.0}&	\textbf{0.0}&	\textbf{0.0}&	\textbf{0.0}&	\textbf{0.0} \\
\bf Poisson &\cite{screened_poisson}   &            1.4&        	1.2&	        5.2&	        3.0&	        28.0&	        7.8&	\textbf{0.0}&	\textbf{0.0}&	\textbf{0.0}&	\textbf{0.0}&	\textbf{0.0}&    \textbf{0.0} \\
\bf Labatut et al. &\cite{Labatut2009a}&   \textbf{1.0}&	\textbf{1.0}&	        2.6&	        1.2&	         4.0&	        2.0&	        0.8&	0.6&	                18.4&	        0.4&	        14.4&	        6.9 \\
\bf Ours &                             &            1.2&	\textbf{1.0}&	\textbf{1.0}&	\textbf{1.0}&	\textbf{1.2}&	\textbf{1.1}&	\textbf{0.0}&	2.0&	0.2&	1.2&	\textbf{0.0}&	0.7\\
\bottomrule
\end{tabular}

\label{tab:berger}
\end{table*}

\begin{figure}[t]
\resizebox{.95\columnwidth}{!}{
\begin{tikzpicture}
\begin{axis}[
    xlabel={Completeness},
    ylabel={Accuracy},
    y label style={at={(axis description cs:0.12,.5)},rotate=0,anchor=south},
    xmin=65, xmax=100,
    ymin=55, ymax=100,
    xtick={65,70,80,90,100},
    ytick={55,60,70,80,90,100},
    legend pos=north west,
    ymajorgrids=true,
    grid style=dashed,
    legend style={font=\small}
]
\addplot[
    dashed,
    color=blue,
    mark=*,
    mark options={solid},
    forget plot
    ]
    coordinates {
(69.6,	75.7)
(79.6,	85.9)
(85.8,	92.5)
(92.1,	97.1)
    };
\addplot[
    color=blue,
    mark=*,
    ]
    coordinates {
(73.1,	71.4)
(82.8,	82.2)
(88.5,	89.6)
(93.8,	95.3)
    };
    \addlegendentry{Ours}
\addplot[
    dashed,
    color=red,
    mark=diamond,
    mark options={solid},
    forget plot
    ]
    coordinates {
(75.9,	65.3)
(85.8,	77.9)
(91.3,	87.4)
(95.6,	94.8)
    };
\addplot[
    color=red,
    mark=diamond,
    ]
    coordinates {
(79.3,	61.9)
(89.5,	74.6)
(94.9,	84.7)
(98.3,	92.9)
    };
    \addlegendentry{Vu \etal}
\addplot[
    dashed,
    color=green,
    mark=square,
    mark options={solid},
    forget plot
    ]
    coordinates {
(75.9,	63.3)
(85.8,	75.9)
(91.3,	85.6)
(95.7,	93.6)
    };
\addplot[
    color=green,
    mark=square,
    ]
    coordinates {
(79.3,	56.9)
(89.5,	69.5)
(94.8,	79.8)
(98.3,	89.3)
    };
    \addlegendentry{Jancosek \etal}
\addplot[
    dashed,
    color=violet,
    mark=triangle,
    mark options={solid},
    forget plot
    ]
    coordinates {
(74.8,	61.8)
(85.8,	74.3)
(92.0,	84.4)
(96.3,	92.5)
    };
\addplot[
    color=violet,
    mark=triangle,
    ]
    coordinates {
    (76.9,	61.0)
    (87.4,	73.5)
    (92.9,	83.4)
    (96.9,	91.6)
    };
    \addlegendentry{Poisson}
\addplot[
    color=black,
    dashed
    ]
    coordinates {
    (0,	0)
    (1,1)
    };
    \addlegendentry{cleaned surface}
\node[] at (axis cs: 73,65) {$5cm$};
\node[] at (axis cs: 98,98) {$50cm$};
\end{axis}
\end{tikzpicture}
}
\caption{\textbf{Performance analysis on ETH3D.}
Each point corresponds to the accuracy and completeness at a given error threshold, respectively at 5, 10, 20 and 50\,cm. Dashed lines represent the performance of meshes cleaned by post-processing. Our method produces meshes with a higher accuracy but a lower completeness.}
\label{fig:eth_precision_recall}
\end{figure}
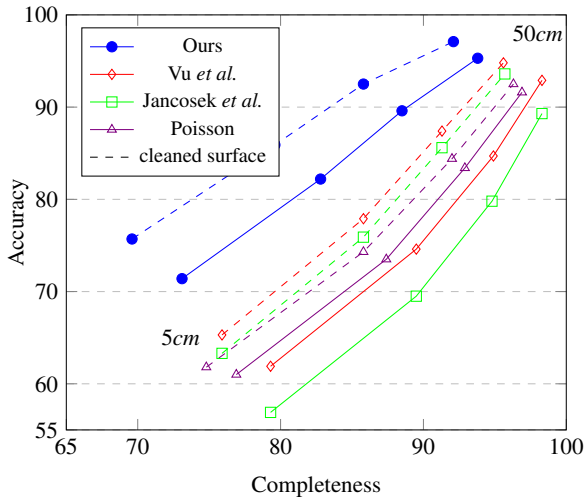

\begin{figure*}[t]
\centering
\includegraphics[width=.95\linewidth]{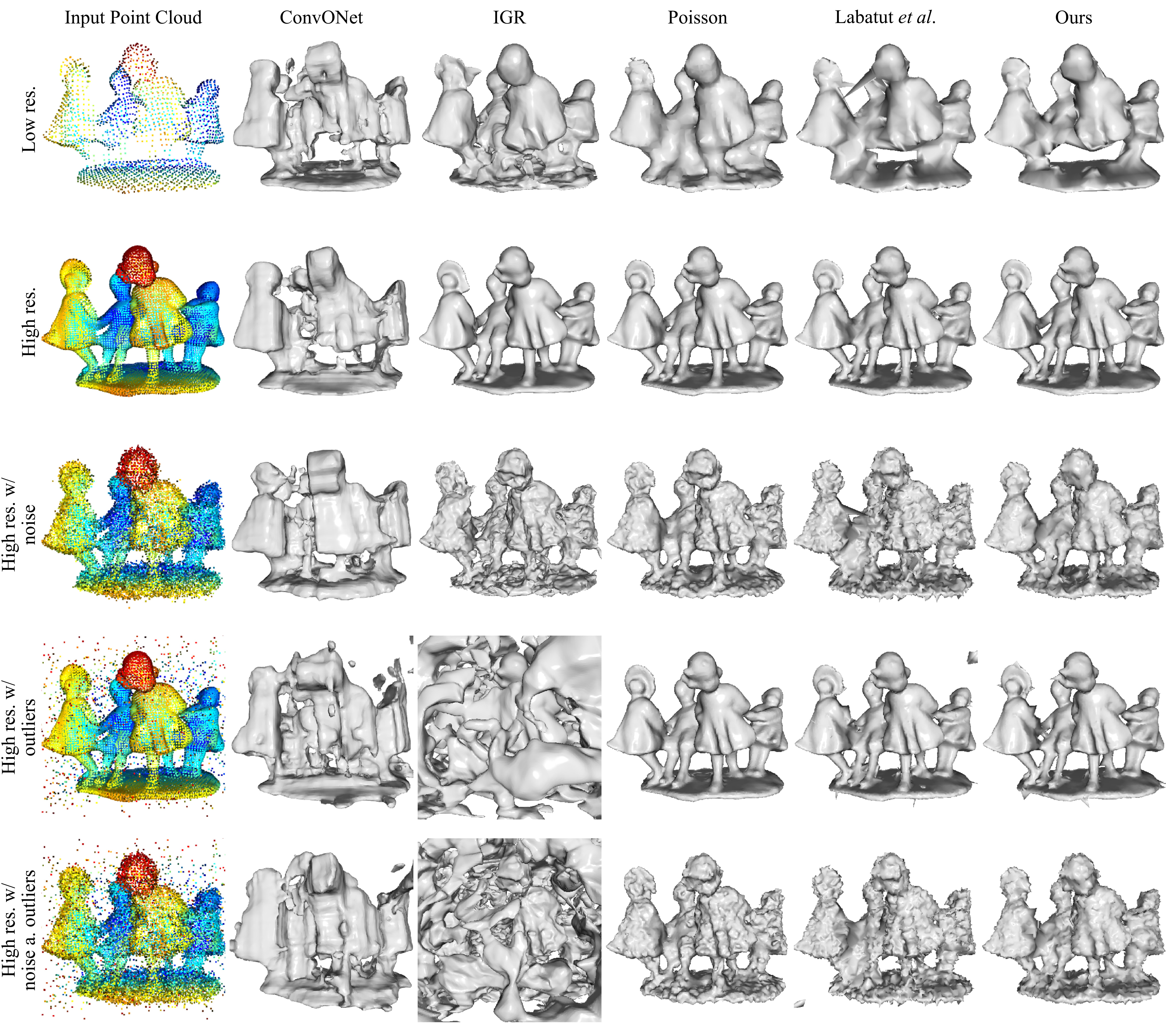}
\caption{
{\textbf{Qualitative Results on Berger et al.'s dataset.} Reconstruction of the \emph{dancing children} shape of Berger et al.'s benchmark \cite{Berger2011} with different levels of noise and outliers to emulate challenging MVS settings. \RAPH{Note that in contrast to ConvONet \cite{Peng2020}, our method generalizes much better to unseen objects, is highly resilient to outliers, and does not produce the floating artifacts of the IGR \cite{Gropp2020} and Labatut \cite{Labatut2009a} algorithms. The Screened Poisson reconstruction \cite{screened_poisson} is visually similar to ours, but occasionally produces unwanted surface parts.}}
}
\label{fig:reconbench}
\end{figure*}
\noindent\textbf{Competing Methods.}
We compare our model with other mesh reconstruction methods that have available (or re-implementable) code, and the ability to scale to large scenes with several million points:
\begin{itemize}[nosep]
    \item \textbf{ConvONet} \cite{Peng2020} is a deep model, like ours, but that does not take visibility into account. We use the \href{https://github.com/autonomousvision/convolutional_occupancy_networks}{model (with multi-plane decoder)} pretrained on the entirety of ShapeNet for the object-level reconstruction. {Among all the available pretrained models, this one gave the best performance.}
    \item \textbf{IGR} \cite{Gropp2020} is a deep model which we retrain for each object using the \href{{https://github.com/amosgropp/IGR}}{official implementation}.
    \item Screened \textbf{Poisson} \cite{screened_poisson} \RAPH{is a classic non-learning-based method which approximates the surface as a level-set of an implicit function estimated from normal and point information. 
    We chose an octree of depth $10$ and Dirichlet boundary condition for object-level {reconstruction} and $15$ and Neumann boundary condition for scene-level reconstruction.}
    \item \textbf{Labatut \etal} \cite{Labatut2009a} is a graph-cut-based method for range scans that makes use of visibility information. We use our own implementation of the algorithm and use the parametrization suggested by the authors ($\alpha_{vis} = 32$ and $\lambda = 5$).
    \item \textbf{Vu \etal} \cite{Vu2012} is an extension of Labatut \etal \cite{Labatut2009a} to \acrshort{mvs} data. We use its OpenMVS \cite{cernea2015openmvs}  implementation with default parameters.
    \item \textbf{Jancosek \etal} \cite{Jancosek2014} also exploits visibility in a graph-cut formulation, with special attention to weakly-supported surfaces. We use the OpenMVS \cite{cernea2015openmvs} implementation with default parameters.
\end{itemize}

For scene reconstruction, we compare all methods with and without mesh cleaning, with the same parameters over all experiments. \RAPH{See the supplementary for details. We also experimented with post-processing the Screened Poisson reconstruction with the included surface trimming tool, but could not find consistent trimming parameters that improve the mean values of Poisson presented in \tabref{tab:berger} and \tabref{tab:results}.}

\subsection{Object-Level Reconstruction}
\label{section:object}
\noindent\textbf{Experimental Setting.}
{To evaluate the adaptability of our method to a wide range of acquisition settings, we use the surface reconstruction benchmark of Berger \etal~\cite{Berger2011}. It includes five different shapes with challenging characteristics such as a non-trivial topology or details of various feature sizes. The provided benchmark software allows to model a variety of range scanning settings to produce shape acquisitions with different defect configurations. We apply to each shape different settings such as varying resolution, noise level, and outlier ratio, meant to reproduce the variety of defects encountered in real-life MVS scans.} 


\noindent\textbf{Results.}
The results are presented in \tabref{tab:berger} and illustrated in \figref{fig:reconbench} (\emph{dancing children} shape) \RAPH{and in the supplementary material} (the other shapes). We observe that the other learning-based methods have a hard time with this dataset. ConvONet~\cite{Peng2020} does not generalize well from the simple models of ShapeNet to the more challenging objects evaluated here. As for IGR \cite{Gropp2020}, it works well in the absence of noise and outliers but produces heavy artifacts on defect-laden point clouds. In contrast, our method is able to generalize to the new unseen shapes and significantly outperforms ConvONet and IGR. Our method also outperforms the state-of-the-art and highly specialized algorithm of Labatut \etal\cite{Labatut2009a}, showing that the graph neural network is able to learn a powerful visibility model with a higher accuracy than methods based only on handcrafted features.

\subsection{Large-Scale Scene Reconstruction}
\label{section:scene}

\begin{figure*}[ht!]
\begin{tabular}{ccccc}
     \begin{subfigure}{.19\linewidth}
       \includegraphics[width=1\linewidth]{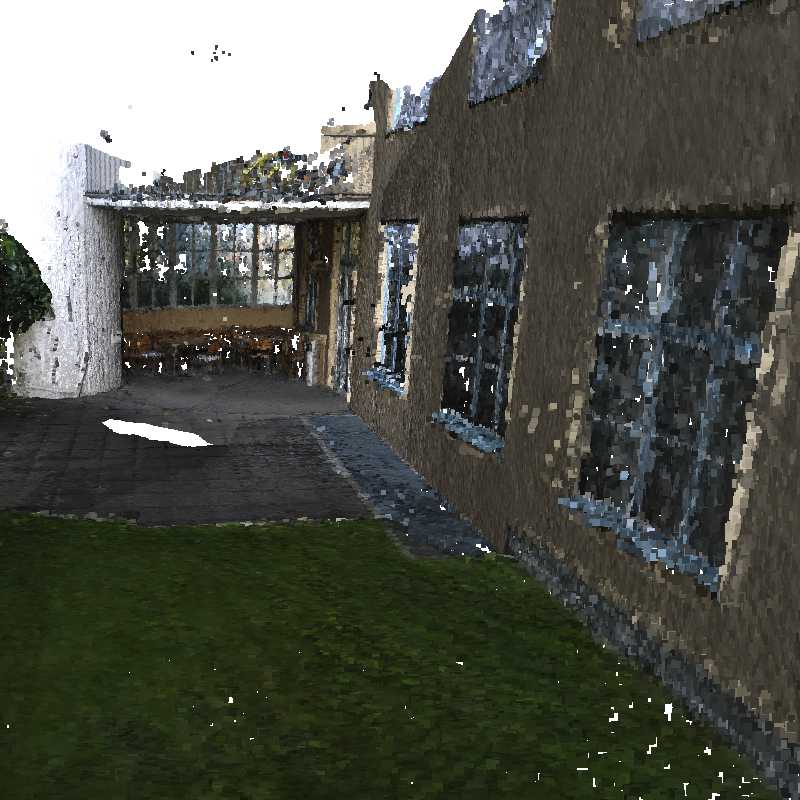}
       \caption{Dense MVS input.}
       \label{fig:teaser1}
     \end{subfigure}
       \begin{subfigure}{.19\linewidth}
       \includegraphics[width=1\linewidth]{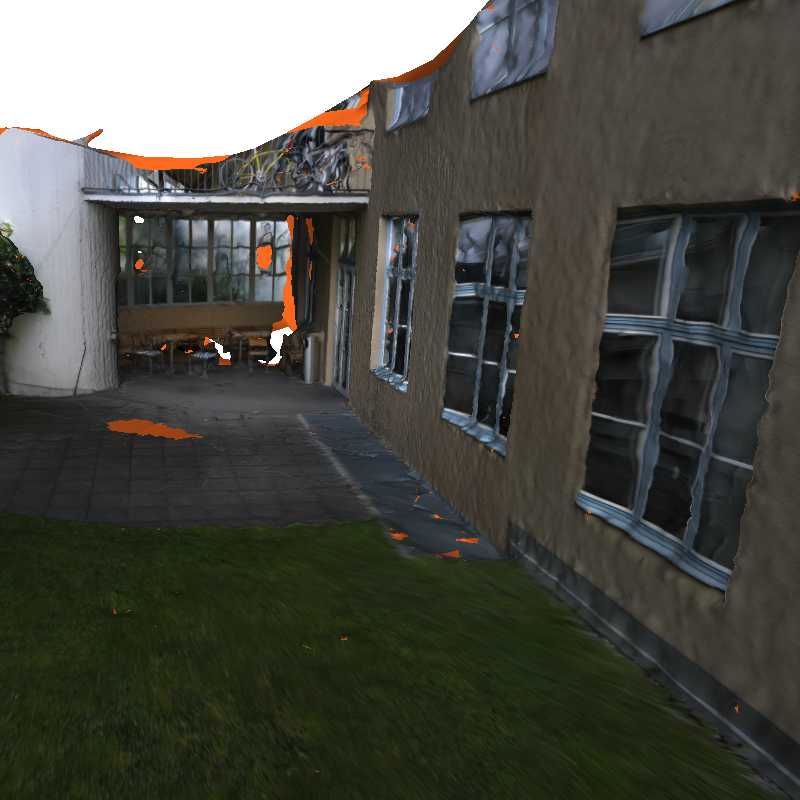}
        \caption{Our textured mesh.}
         \label{fig:teaser2}
        \end{subfigure}
       \begin{subfigure}{.19\linewidth}
       \includegraphics[width=1\linewidth]{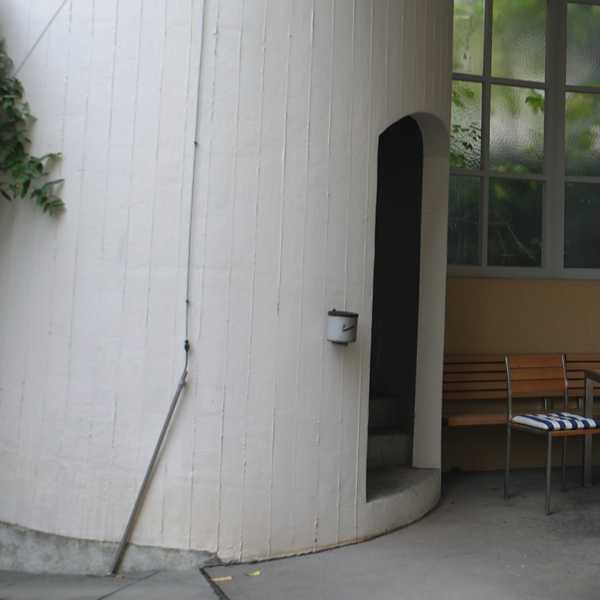}
        \caption{Details Image.}
         \label{fig:teaser3}
        \end{subfigure}    
               \begin{subfigure}{.19\linewidth}
       \includegraphics[width=1\linewidth]{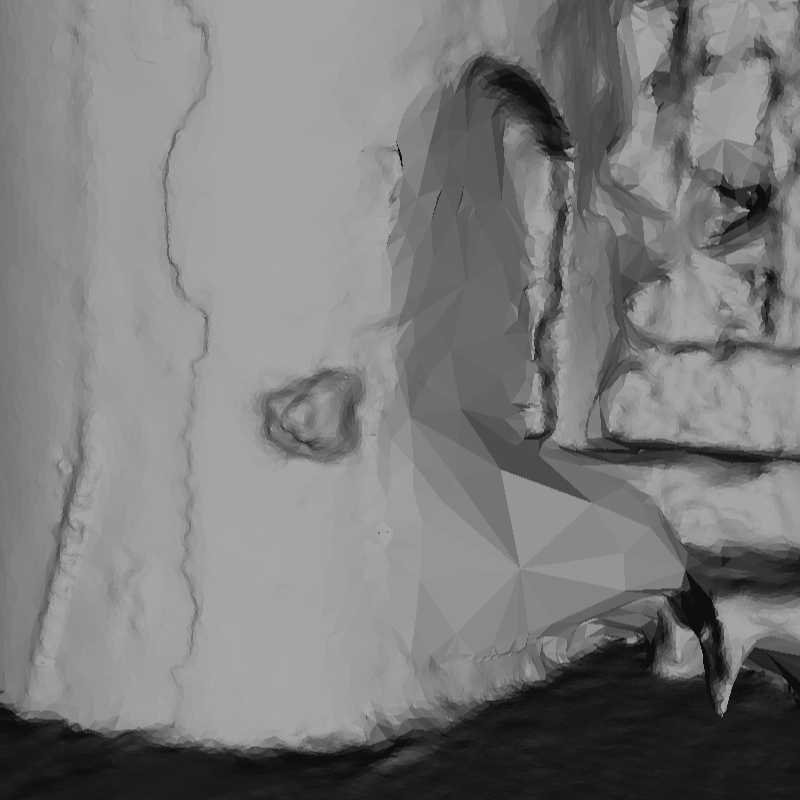}
        \caption{\cite{Jancosek2014}.}
         \label{fig:teaser4}
        \end{subfigure}    
               \begin{subfigure}{.19\linewidth}
       \includegraphics[width=1\linewidth]{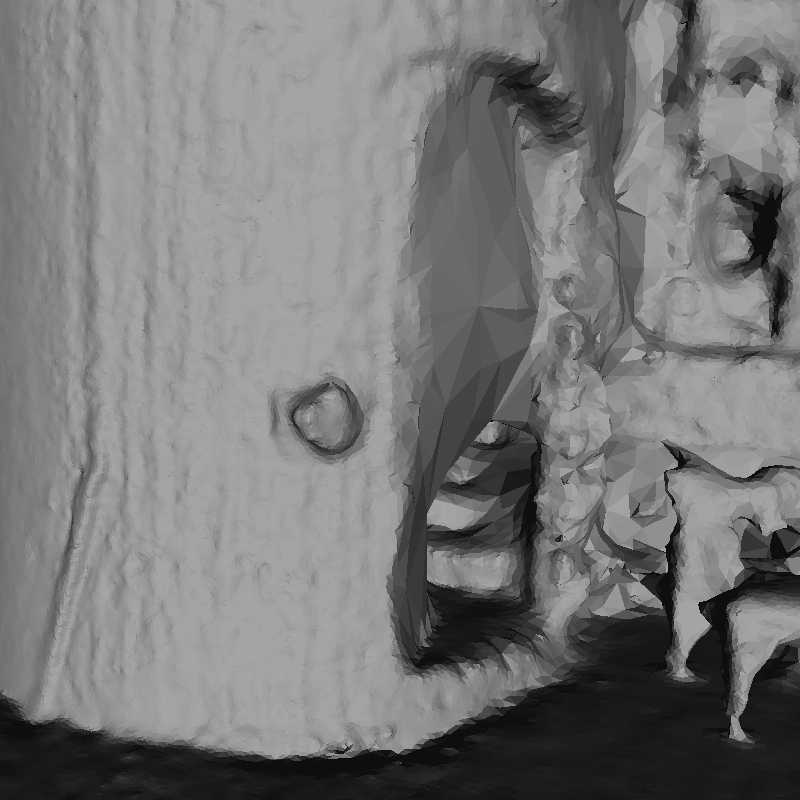}
        \caption{Ours.}
         \label{fig:teaser5}
        \end{subfigure}
\end{tabular}
\caption{\textbf{Qualitative Results on ETH3D.}
Our mesh reconstruction method takes as input a dense MVS point cloud \Subref{fig:teaser1} and produces a mesh~\Subref{fig:teaser2}, simultaneously preserving fine details and completing missing parts (here textured with \cite{lettherebecolor}).
We represent: in \Subref{fig:teaser3}, a cropped image of a detail from the \emph{terrace} scene of the ETH3D benchmark \cite{Schops2017}; in \Subref{fig:teaser4}, the reconstruction by Jancosek and \etal\cite{Jancosek2014}; and 
in \Subref{fig:teaser5}, our reconstruction.
Notice the missing staircase and spurious vertical pattern on the concrete wall 
in \Subref{fig:teaser4}. In contrast, our method \Subref{fig:teaser5} reconstructs part of the staircase as well as the fine-grained wall textures.}
\label{fig:qualeth3D}
\end{figure*}



\noindent\textbf{Experimental Setting.}
To evaluate the ability of our method to scale to entire scenes, we experiment with the high-resolution \acrshort{mvs} benchmark ETH3D \cite{Schops2017}. This benchmark is originally designed to evaluate \acrshort{mvs} algorithms (point cloud reconstruction) under challenging real-life conditions. Ground-truth point clouds and camera poses are openly available for a training set including $7$ indoor and $6$ outdoor scenes. The ground truth consists of LiDAR scans post-processed to only contain reliable points.

While we cannot train our network on this dataset due to the lack of closed surfaces in the ground truth, we can evaluate the quality of the output of our algorithm after sampling points on the reconstructed surface. To this end, we generate dense point clouds from downsampled images ($3100\times2050$ pixels) of the $13$ training scenes using a patch-based \acrshort{mvs} algorithm \cite{patch_match} implemented in OpenMVS \cite{cernea2015openmvs}. The point clouds and associated camera poses are used as inputs for all mesh reconstruction methods evaluated in \secref{sec:evaluation}. Additionally, as input for the Screened Poisson and IGR algorithm, we estimate surface normals using Jets \cite{jet-normal} and consistently orient them towards the sensor.

We also assess the scalability and generalization capability of ConvONet on real world outdoor scenes. We use the volume decoder model pretrained on the synthetic indoor scene dataset \cite{Peng2020} operating on a sliding window. To avoid prohibitively expensive computations caused by far away outliers, we manually crop most of the scenes to limit the bounding volume. It is important to note that our method requires no such preprocessing. However, even in this prepared setting, the resulting surfaces were of significantly lower quality than all other methods. This can be explained by the fact that, even if the ConvONet model was trained on \RAPH{collections of ShapeNet models}, the distribution of objects in this training set is very different from the real-life scenes \RAPH{of ETH3D}. Our model being purely local, does not suffer from this lack of generalizability. \RAPH{See the supplementary for a qualitative comparision of ConvONet and our method.} A time and memory comparison between ConvONet, Vu \etal and our method is given in \tabref{tab:memory}. As for IGR, the size of its network prevents us from reconstructing ETH3D scenes. Consequently, in the following, we exclude ConvONet and IGR from evaluations on ETH3D.

\begin{table}
\caption{\textbf{Quantitative results on ETH3D.} We report the following extrinsic and intrinsic metrics on the ETH3D dataset: F1-Score at 5\,$cm$ (F1), number of connected components (CC), and surface area of the mesh in square meters $\times 10^{-2}$ (Area). Numbers in parentheses are from the meshes before the cleaning step.}
\begin{center}
\begin{tabular}{lccc}
\hline
\textbf{Method} & \textbf{F1} & \textbf{CC} & \textbf{Area} \\\hline 
Poisson \cite{screened_poisson} & 66.8 (67.2) & 83 (23131) & 82 (116) \\
Vu \etal \cite{Vu2012}              &  70.6 (70.8)  & 17 (560) & 17 (125) \\
Jan. \etal \cite{Jancosek2014}  & 70.0 (67.7)  & 14 (667) & 14 (78) \\ 
Ours                                &  \bf73.1 (72.1)  & 23 (253) & 11 (24) \\
\end{tabular}
\end{center}
\label{tab:results}
\end{table}

\noindent\textbf{Evaluation.}
We use the ETH3D \cite{Schops2017} multi-view evaluation procedure. This protocol accounts for incomplete ground truth by segmenting the evaluation space into \emph{occupied}, \emph{free} and \emph{unobserved} regions. As the procedure was designed to evaluate point reconstruction, we uniformly sample random points from the reconstructed meshes. We then evaluate the methods on extrinsic quality parameters, namely (i) accuracy, (ii) completeness, and (iii) F1-score (harmonic mean). Accuracy is defined as the fraction of sampled points on the output mesh within distance $\tau$ to any point in the ground truth. Reconstructed points in the unobserved space are thereby ignored. Completeness is defined as the fraction of ground-truth points for which there exists a point sampled on the reconstructed mesh within a distance~$\tau$.

It is important to note that the ETH3D stereo benchmark is typically used to evaluate the quality of point clouds produced by dense MVS methods. In contrast, the approaches we evaluate concern 
the reconstruction of compact and watertight meshes. It is a much harder task. Watertightness {in particular,} requires special attention regarding holes and close parallel surfaces, while algorithms producing point clouds may ignore {such considerations}. Consequently, the comparison of the methods we evaluate with other entries of the ETH3D benchmark is not valid.

\noindent\textbf{Results.} In \figref{fig:eth_precision_recall}, we present the accuracy-completeness curve for $\tau = 5,~10,~20~\text{and}~50$\,cm, illustrating the varying trade-offs between completeness and accuracy for the different methods. For instance, at $\tau = 50$\,cm, all methods have an F1-score of around $95\%$. In comparison, our method provides a lower completeness but a higher accuracy; nevertheless, it results in a better overall F-score. The higher accuracy provided by our method is illustrated in \figref{fig:teaser} and \figref{fig:qualeth3D}, where fine details that are hard to reconstruct are better preserved. In \tabref{tab:results}, we report intrinsic mesh quality measures at $\tau=5$\,cm for different methods. We improve the F-score by $2.5$ points, while producing a surface up to $35\%$ more compact. More results can be found in the appendix.

We would like to stress that, while our model is learning-based, its training set \cite{shapenet2015} is very different from the one used to evaluate the performance \cite{Schops2017}: {we train {our method} on few artificial, simple and closed objects, while we evaluate on complex real-life scenes.
Furthermore, our network does not optimize towards the main evaluation metrics. Instead, we optimize towards a high volumetric IoU of outside and inside cells. This implies that our model, while being simple, can learn a {relevant} visibility model that is able to generalize to data of unseen nature.}

\noindent\textbf{Speed and Memory.}
{In \tabref{tab:memory}, we report the speed and GPU memory requirements of the different competing methods. Our approach compares favorably to ConvONet for all space-time trade-offs, on top of improved reconstruction metrics. While the added \acrshort{gnn} inference step of our method results in a slower overall prediction compared to \cite{Vu2012}, we argue that the added accuracy justifies the extra processing time. Our method can process the entirety of the ETH benchmark in under $25$ minutes.}

{Besides, thanks to our efficient modified GraphSAGE scheme, the unary potentials can be computed purely locally; global prediction agreement is achieved by the graph cut. 
We can control precisely the memory usage by choosing the number of tetrahedra to process at a time, each one using around $10$\,MB of memory. This memory usage can be further improved with a memory-sharing scheme between nodes, allowing us to process up to $400,000$ tetrahedra simultaneously with $8$\,GB of VRAM, which is the same amount of memory necessary for ConvONet to process a single sliding window.}

\begin{table}[t]
\centering
\caption{\textbf{Time and memory footprint.}  We report, for the reconstruction of the \emph{meadow} scene (ETH3D), the computation time for point/tetrahedron features (Feat), tetrahedralization (3DT), network inference (Inference), graph cut (GC), and marching cubes (MC).  {Batch size is given in number of subgraphs\,/\,sliding windows.} Our model alone fills 470\,MB of VRAM, while ConvONet fills 540\,MB.
}
\resizebox{\columnwidth}{!}{%
\begin{tabular}{@{}l@{~}cccc@{~~}ccc@{}}
\toprule
\textbf{} & \textbf{\llap{Batch} size}\!\!\! & \textbf{Feat.} & \!\!\!\textbf{3DT}\!\!\!\!\!\! & \multicolumn{2}{c}{\textbf{Inference}} & \!\!\!\!\!\!\textbf{GC\,/\,MC}\!\!\! & \textbf{Total} \\ \midrule
Vu et al. \cite{Vu2012} & -  & 13\,s & 4\,s  & - & -           & 14\,s & \textbf{31\,s}  \\
Ours     & 400k & 14\,s & 4\,s  & 24\,s & 7.9 GB & 16\,s & 58\,s \\
Ours     & 1& 14\,s & 4\,s  &  75\,s & \textbf{0.5 GB}   & 16\,s & 109\,s \\ 
ConvONet \cite{Peng2020} & 1   & 5\,s  & - & 145\,s & 7.9 GB & 14\,s & 164\,s\\
\bottomrule
\end{tabular}
}
\label{tab:memory}
\end{table}
\subsection{Design Choices and Ablation Study}
\label{sec:ablation}
In this section, we evaluate the effect of several of our design choices on the performance of our algorithm.

\noindent\textbf{Direct Prediction.} We assessed the impact of the graph cut step by evaluating the quality of the surface obtained using only the unary terms: tetrahedrons with an insideness over $0.5$ are predicted as inside, and the others outside.
This leads to very fragmented reconstructed surfaces (over $10$ times more components), especially in the background of the scenes.
Given that our objective is to produce compact watertight surfaces, we chose to use a regularization, here with a global energy minimization. 


\noindent\textbf{Learning Binary Weights.}
{We designed an \acrshort{gnn} able to predict binary weights in the energy model along the unaries. However, this lead to more fragmented surfaces and overall lower performance. The difficulties of learning the potentials of an energy model with a neural network are expected, as 
neural networks operate locally and in continuous space, while graph cuts operate globally and in discrete space.
In fact, we can interpret our \acrshort{gnn} prediction as the marginal posterior inside/outside probability of each tetrahedron, while the graph cut provides an inside/outside labeling of maximum posterior likelihood (MAP) in a fitting Potts model \cite{boykov2001fast}.
These two tasks being conceptually different, we were not able to successfully learn our surface reconstruction in an end-to-end fashion and leave this endeavor for future work. }

\noindent\textbf{Graph Convolution.} We tried replacing our GNN scheme with the Dynamic Edge Conditioned Convolution of Simonovsky \etal~\cite{Simonovsky2017} for its ability to leverage facet features derived from both ray and tetrahedrons. {This resulted in a marginal increase in performance (under $1\%$ decrease of the Chamfer distance) at the cost of an increase in computational and memory requirements. For the sake of simplicity and with scalability in mind we keep the simple GraphSAGE scheme.}   

\noindent\textbf{Relevance of Geometric Features.} We tried training a model using only visibility features and no tetrahedron-level geometric features. In doing this ablation, we lose between 10\% of F1-Score on ETH3D. This demonstrates that visibility information should be combined with geometric information, which is not typically done in traditional approaches. 

\noindent\LOIC{\textbf{Limitations.}
As learning-based methods in general, our approach requires the training and test datasets to have comparable distributions. However, since the inference is purely local, we do not need both datasets to contain similar objects. Yet the characteristics of the acquisition must be similar in terms of accuracy and density.}
\LOIC{Besides, as common in Delaunay-based methods, our reconstructed surface is bound to go through the triangles of the Delaunay tetrahedralization. This can limit precision when the acquisition is noisy, and prevent us from reconstructing details below the sampling resolution.}

\section{Conclusion}

{We propose a scalable surface reconstruction algorithm based on graph neural networks and graph-cut optimization. Our method, trained from a  small artificial dataset, is able to rival with state-of-the-art methods for large-scale reconstruction on real-life scans.
Thanks to the locality of the prediction of the unary potentials associated with tetrahedra, our method can perform inference on large clouds with millions of tetrahedra.
Our approach demonstrates that it is possible for deep-learning techniques to successfully tackle hard problems of computational geometry at a large scale.
}

\section*{Acknowledgments} 

This work was partially funded by the ANR-17-CE23-0003 BIOM grant.
\FloatBarrier
{\small
\balance
\bibliographystyle{eg-alpha-doi} 
\bibliography{bibliography}
}
\newpage
\twocolumn[{
\begin{center}
     \huge \textbf{Supplementary Material for:\\
     Scalable Surface Reconstruction with\\ Delaunay-Graph Neural Networks}
\end{center}
~\\~\\
}]

In this supplementary document, we first provide additional information about our our training data in \secref{sec:training} and implementation in \secref{sec:implementation}. Finally, we provide additional qualitative and quantitative experimental results in \secref{sec:object} for object-level reconstruction, and in \secref{sec:scene} for scene-level reconstruction.

\section{Generating Training Data}
\label{sec:training}
\label{generation_of_training_data}
In an ideal setting, we would have trained our network on real-life, large-scale, \acrshort{mvs} acquisitions together with associated ground-truth surfaces. However, such surfaces are difficult to produce. Two methods can be used to circumvent this issue: using laser scans or resorting to synthetic scans.

\noindent\textbf{Laser Scans.}
The first option is to use a surface reconstructed from a high-precision acquisition of a scene, e.g., with a stationary LiDAR scan. In parallel, the scene can be captured by cameras to produce an \acrshort{mvs} acquisition, typically of lower quality. This procedure has been used in several \acrshort{mvs} benchmarks \cite{Strecha08, Knapitsch2017, Schops2017, Seitz2001}. However, a difficulty remains when reconstructing the ground-truth surface. We require a closed surface to derive the ground-truth occupancy. The chosen surface reconstruction method may introduce biases in the ground-truth surface, such as over-smoothing. Additionally, even with high-quality LiDAR acquisitions, parts of the scene can be missing, e.g., due to occlusions. These issues ultimately lead to inconsistencies in the training data, because the \acrshort{mvs} acquisition locally diverges from the ground-truth surface. Thus, in practice, we found that the incompleteness of available LiDAR scans makes this source of data too unreliable to train our network.



\noindent\textbf{Synthetic Scans.} A second option for producing ground-truth data is to use synthetic scans of closed artificial shapes. To this end, we make use of the range scanning procedure from the Berger \etal \cite{Berger2011} benchmark for surface reconstruction.

We modified 
the provided code to export the camera positions of the scanning process.
We then synthetically scan artificial shapes using our modified version of the Berger \etal scanning software. We choose at random one of the 5 scanner settings described in \tabref{tab:berger_confs} to scan each training shape. 
The low resolution scanner setting produces uniform point clouds, similar to those obtained by coarse voxelizations. High resolution settings produce point clouds similar to those obtained by MVS. We also add outliers to the scans in the form of randomly distributed points in the bounding box of the objects and associate these points with a random camera position. 
We use this method to produce training data from a small subset of 10 shapes of each of the 13 classes of the ShapeNet subset from \cite{choy20163dr2n2}. We produce watertight meshes of the ShapeNet models using the method of Huang \etal \cite{huang2018shapenetwatertight}. 

\begin{figure*}[t]
\centering
\includegraphics[width=\linewidth]{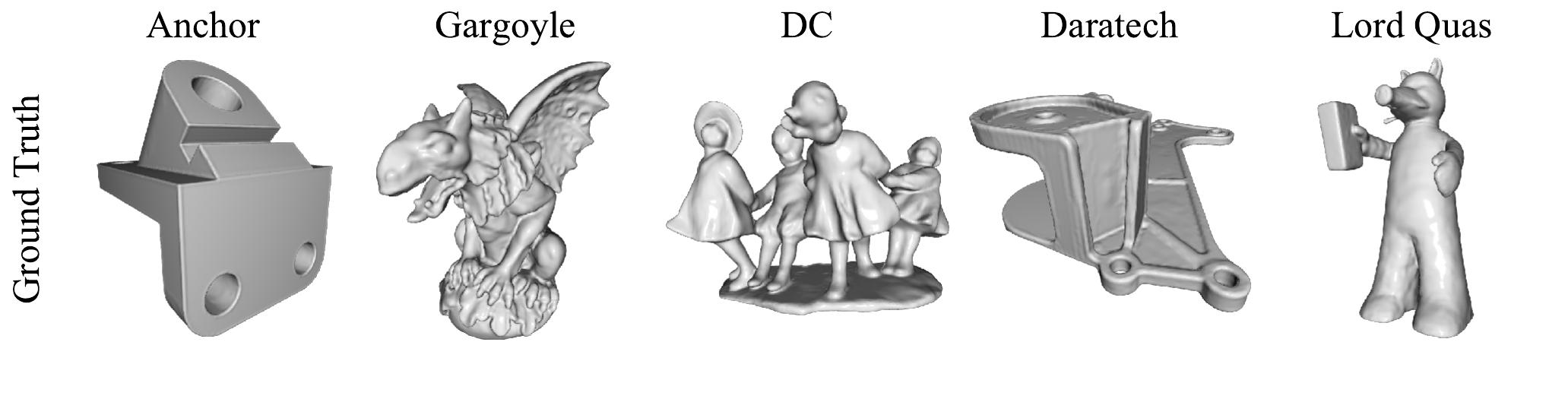}
\caption{\textbf{Ground-truth meshes for Berger et al.'s benchmark.} We represent the $5$ shapes chosen from the Berger et al.'s benchmark \cite{Berger2011} for our evaluation.
}
\label{fig:reconbench_gt}
\end{figure*}

{To obtain the ground-truth occupancy, we sample 100 points in each tetrahedron and determine the percentage of these sampled points lying inside their corresponding ground-truth models. In total, we train our network on around 10M tetrahedra. We also apply the scanning procedure with the 5 different configurations to each shape of the 5 ground-truth shapes from the Berger \etal \cite{Berger2011} benchmark.} See \figref{fig:reconbench_gt} for the 5 ground-truth shapes and the first column of Figures \ref{fig:anchor}-\ref{fig:lordquas} for their scans.
We refer the reader to the original benchmark paper \cite{Berger2011} for further details about the scanning process.



\begin{table*}[]
\caption{\textbf{Scanning configuration for Berger et al.'s benchmark.} We show the five different scanner configurations used in our modified version of the Berger et al.'s scanning procedure. We use the resulting scans to evaluate object-level reconstruction with varying point-cloud defects and for training data generation. For the low resolution (LR) scans the scanning process results in $1000$ to $3000$ points per shape, and for the high resolution (HR), the scanning process yields around $10\,000$ to $30\,000$ points.}
\resizebox{\textwidth}{!}{%
\begin{tabular}{@{}lccccc@{}}
\toprule
\multicolumn{1}{l}{} & \textbf{Low res. (LR)} & \textbf{High res. (HR)} & \textbf{HR + noise (HRN)} & \textbf{HR + outliers (HRO)} & \textbf{HR + noise + outliers (HRNO)} \\ \midrule
Camera resolution x, y & 50, 50 & 100, 100 & 100, 100 & 100, 100 & 100, 100 \\
Scanner positions & 5 & 10 & 10 & 10 & 10 \\
Min/max range & 70/300 & 70/300 & 70/300 & 70/300 & 70/300 \\
Additive noise & 0 & 0 & 0.5 & 0 & 0.5 \\
Outliers (\%) & 0 & 0 & 0 & 0.1 & 0.1 \\ \bottomrule
\end{tabular}%
}
\label{tab:berger_confs}
\end{table*}

\section{Implementation Details}

\label{sec:implementation}

\noindent\textbf{Multi-View Stereo.}
Our implementation relies on the \href{http://cdcseacave.github.io/openMVS/}{OpenMVS} \cite{cernea2015openmvs} library for many of the \acrshort{mvs} processing steps.

We generate dense point clouds using the provided camera poses of all scenes of the ETH3D test dataset. We use the DensifyPointCloud tool of OpenMVS with standard settings, except for the following parameters: \emph{number-views-fuse = 2}, \emph{optimize = 0} and \emph{resolution-level = 4}. 

\noindent\textbf{Visibility-augmented 3DT.}
We use \href{https://www.cgal.org/}{CGAL} to obtain the Delaunay Triangulation and for ray tracing. 
For the ray tracing, we only use one camera per point. We chose  the camera minimizing the angle between the line-of-sight and the point's normal (obtained by local principal component analysis). In our experiments, this allows for a significant speed-up in the ray tracing step with a negligible difference on the predicted surface. Likewise, we disregard the third tetrahedron encountered after a line of sight traverses an observed point,
and beyond (see Fig.~\ref{fig:ray_features}). 

\noindent\textbf{Deep Learning.}
Finally, we use {PyTorch} \cite{paszke2019pytorch} and PyTorch Geometric \cite{fey2019fast} for implementing the graph neural network training and inference.

\noindent\textbf{Binary Weights.}
We use the same surface quality term $B_{s,t}(i_s,i_t) = \mathds{1}(i_s \neq i_t) \,\beta_{s,t}$ as Labatut \etal\cite{Labatut2009a} for a facet interfacing the tetrahedra $s$ and $t$. 
%
Considering the intersection of the circumspheres of $s$ and $t$ with the facet, with angles $\phi$ and $\psi$, then $\beta_{s,t}$ is  defined as:
\begin{align}
\label{eq:beta_skeleton}
  \beta_{s,t} & = 1 - \min \{cos(\phi), cos(\psi)\} ~.
\end{align}

\noindent\textbf{Parameterization of Competing Methods.}
We use the OpenMVS implementations of Vu \etal and Jancosek \etal through the ReconstructMesh tool with \emph{min-point-distance = 0.0}.
For Vu \etal we set \emph{free-space-support = 0}, and we set it to $1$ for Jancosek \etal. 

\RAPH{For the reconstructions of ConvONet we use the multi-plane decoder model pretrained on ShapeNet for object-level reconstruction and the volume decoder model pretrained on the synthetic indoor scene dataset \cite{Peng2020} for scene-level reconstruction, where we set the voxel size to $4~cm$.}

\noindent\textbf{Cleaning of scene reconstruction.} We use default clean options in OpenMVS for the cleaning step for all scene-level mesh reconstructions.

\section{Object-Level Reconstruction}
\label{sec:object}

\noindent\textbf{Metrics.}
We evaluate object-level reconstruction with the volumetric IoU, the symmetric Chamfer distance, the number of connected components and the number of non-manifold edges in the reconstructed mesh.


For the Chamfer distance, we sample $n_{S} = 100\,000$ points on the ground-truth meshes $\mathcal{M_G}$ and reconstructed meshes $\mathcal{M_P}$. The distances between the resulting ground-truth point cloud $S_G$ and the reconstruction point cloud $S_P$, approximating the two-sided Chamfer distance, is then given as:

\begin{align}\nonumber
    d_{CD}(\mathcal{M_G},\mathcal{M_P}) =
    &\frac{1}{n_S} \sum_{x \in S_G} \min_{y \in S_P} {\vert \vert x - y \vert \vert }_2^2 \\
    + &\frac{1}{n_S}\sum_{y \in S_P} \min_{x \in S_G} {\vert \vert y - x \vert \vert }_2^2
    \label{eq:chamfer}
\end{align}

The volumetric IoU is defined as:

\begin{align}
    \text{IoU}(\mathcal{M_G},\mathcal{M_P}) =
    &\frac{\vert \mathcal{M_G} \cap \mathcal{M_P} \vert}{\vert \mathcal{M_G} \cup \mathcal{M_P} \vert},
    \label{eq:iou}
\end{align}

We approximate the volumetric IoU by sampling $100\,000$ points in the union of the bounding boxes of the ground-truth and reconstruction meshes.




For the number of connected components, we count all components of the reconstructed meshes. The ground-truth meshes all have only one component. Additionally, they do not have any non-manifold edges.

\begin{figure*}[p]
\centering
\includegraphics[width=1\linewidth]{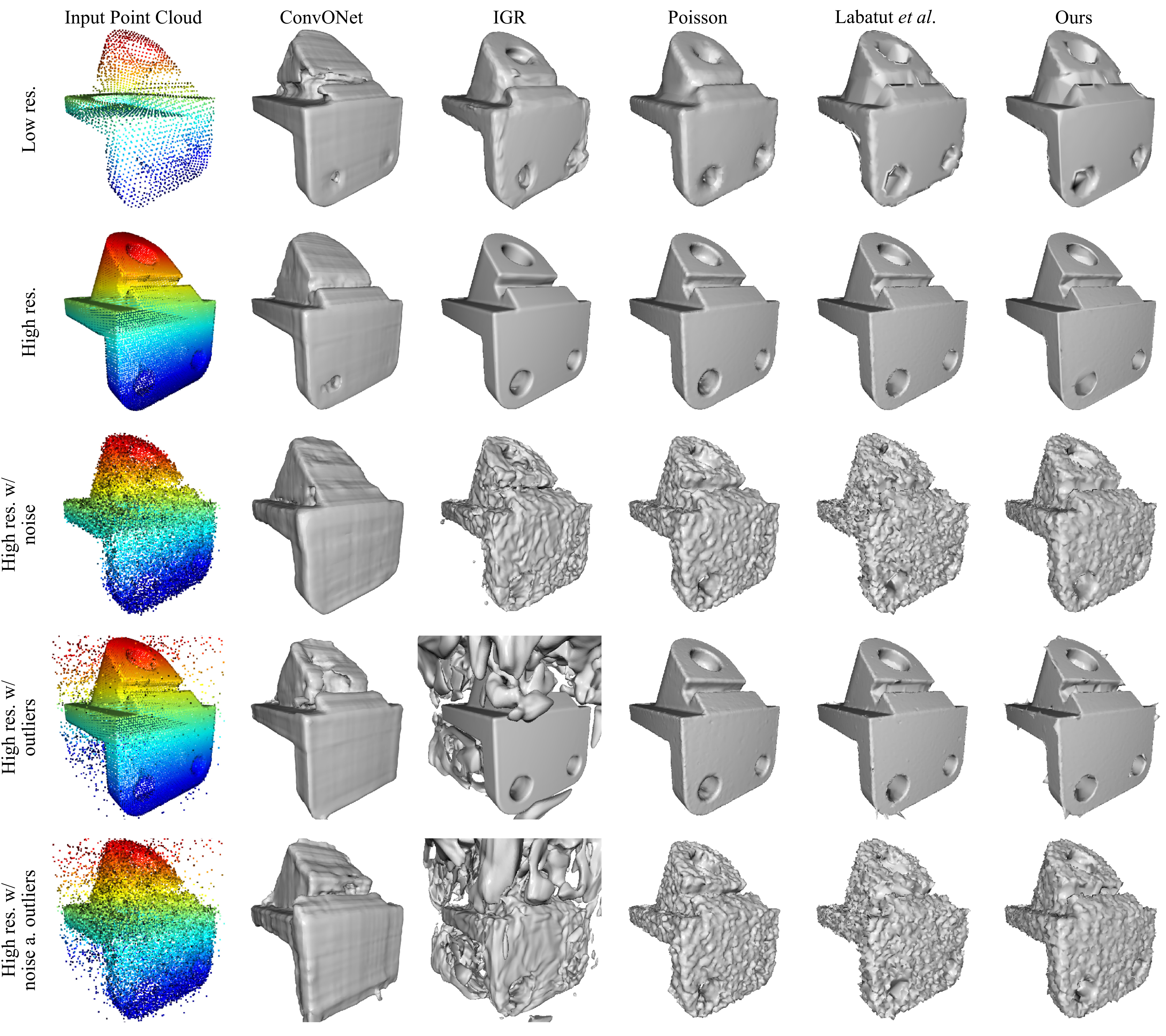}
\caption{\textbf{Reconstruction of the \emph{Anchor} object in the surface reconstruction benchmark of Berger \etal\ \cite{Berger2011}.} We show the input point clouds in column~1. ConvONet \cite{Peng2020} (column~2) does not generalize well to the unseen new shape. IGR \cite{Gropp2020} (column~3) works well at high resolution but fails in the other cases. The Screened Poisson \cite{screened_poisson} algorithm (column~4)  does not reconstruct the sharp features well, but is robust against outliers, even close to the surface. The reconstructions of Labatut \etal \cite{Labatut2009a} (column~5) and ours (column~6) are visually similar for the easier high resolution case. Our method performs slightly better on the low resolution, and noise cases.
}
\label{fig:anchor}
\end{figure*}

\begin{figure*}[p]
\centering
\includegraphics[width=1\linewidth]{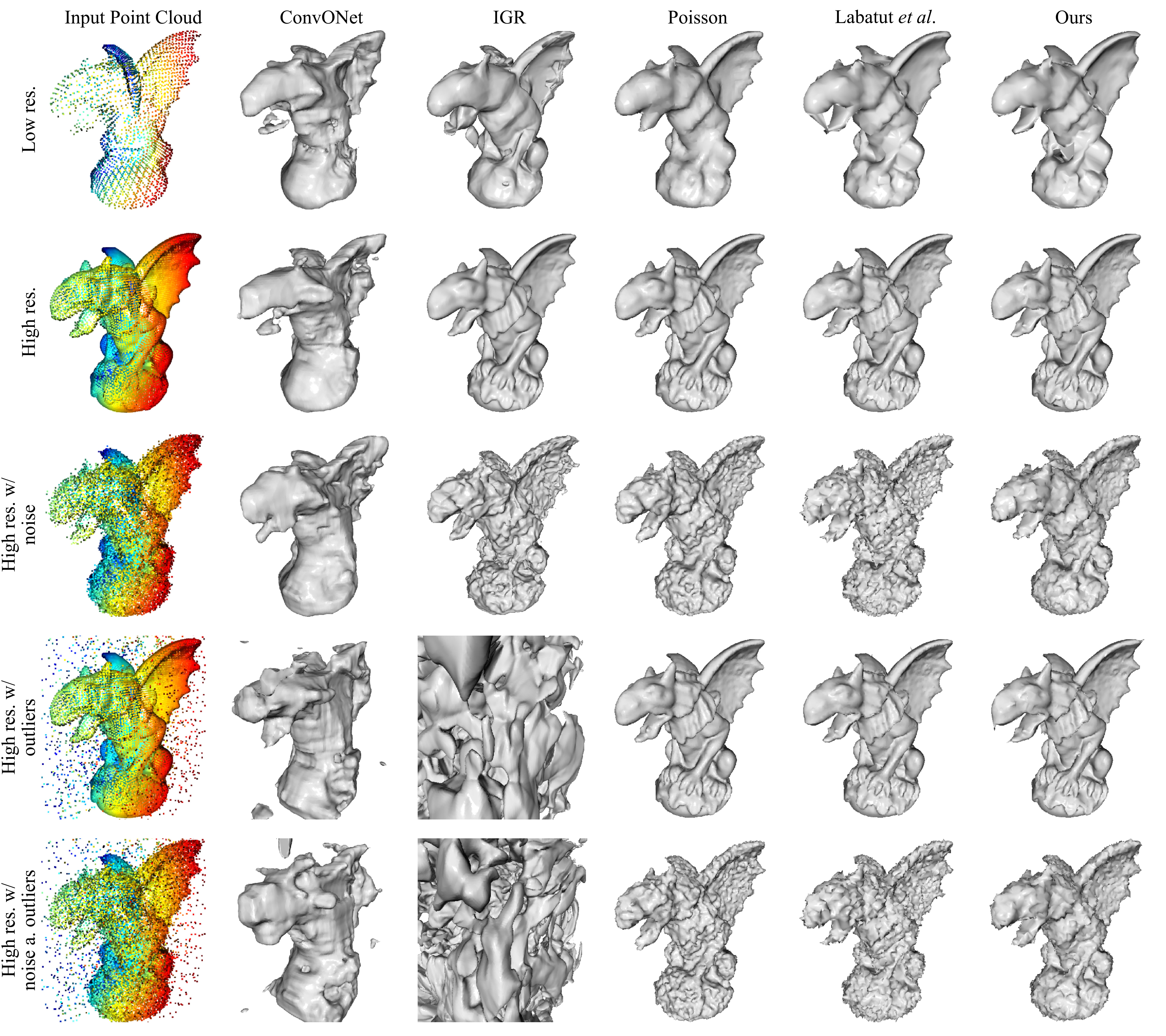}
\caption{\textbf{Reconstruction of the \emph{Gargoyle} object in the surface reconstruction benchmark of Berger \etal\ \cite{Berger2011}.} We show the input point clouds in column~1. ConvONet \cite{Peng2020} (column~2) does not generalize well to the unseen new shape. IGR \cite{Gropp2020} (column~3) generates many surface components from outliers. The Screened Poisson \cite{screened_poisson} algorithm (column~4) does not reconstruct the sharp features well, but is robust against outliers, even close to the surface. The reconstructions of Labatut \etal \cite{Labatut2009a} (column~5) and ours (column~6) are visually similar for the easier high resolution case. While both methods are very robust against outliers, our method performs slightly better on the low resolution, outlier and noise cases.}
\label{fig:gargoyle}
\end{figure*}

\begin{figure*}[p]
\centering
\includegraphics[width=1\linewidth]{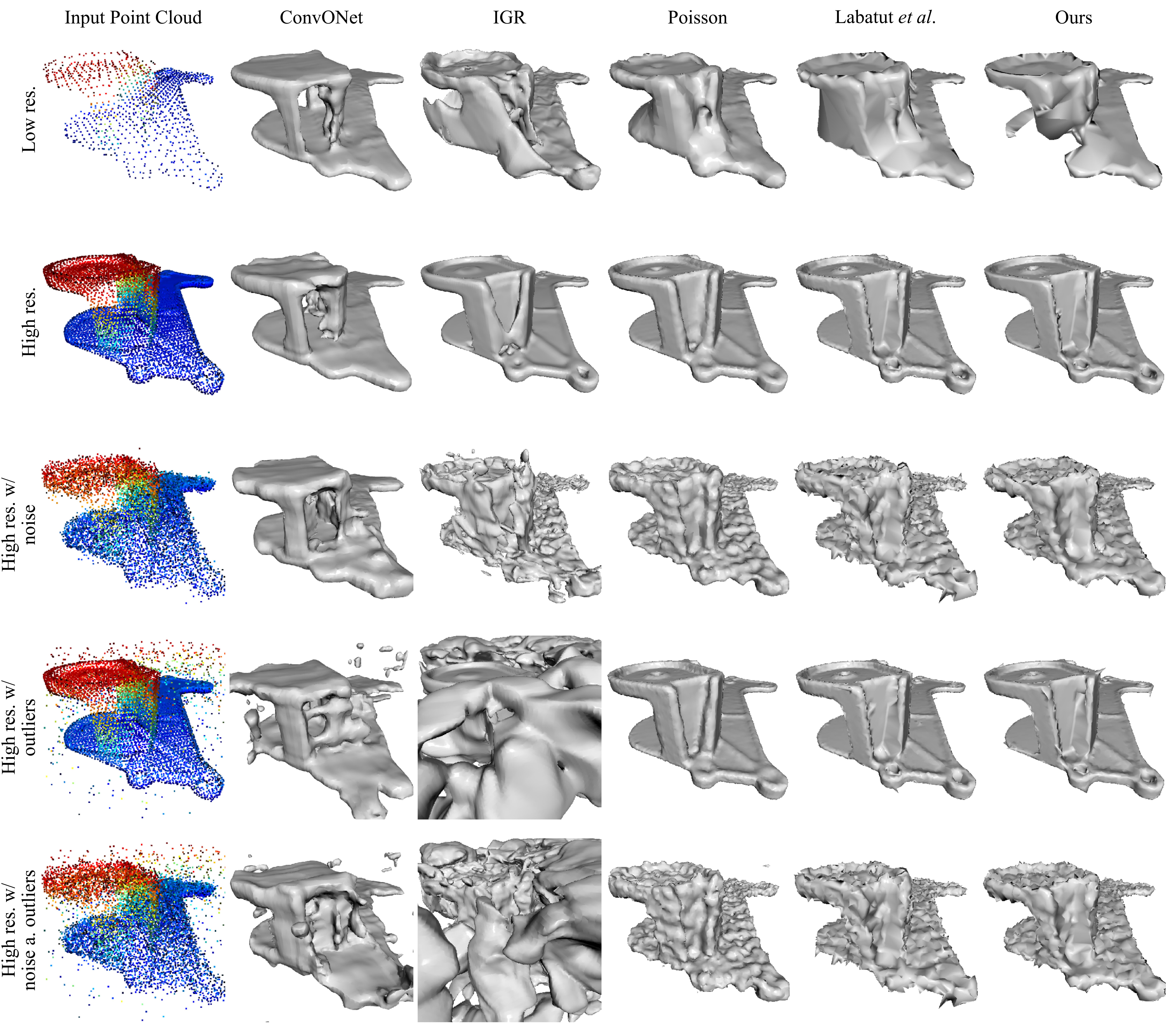}
\caption{\textbf{Reconstruction of the \emph{Daratech} object in the surface reconstruction benchmark of Berger \etal\ \cite{Berger2011}.} We show the input point clouds in column~1. ConvONet \cite{Peng2020} (column~2) does not generalize well to the unseen new shape. 
As with other shapes, IGR \cite{Gropp2020} (column~3) works well at high resolution but generates artefacts or fails in other settings.
The Screened Poisson \cite{screened_poisson} algorithm (column~4) does not reconstruct the sharp features well, but is robust against outliers, even close to the surface. In the low resolution setting, our algorithm is incomplete where Labatut creates unwanted surface parts.
}
\label{fig:daratech}
\end{figure*}

\begin{figure*}[p]
\centering
\includegraphics[width=1\linewidth]{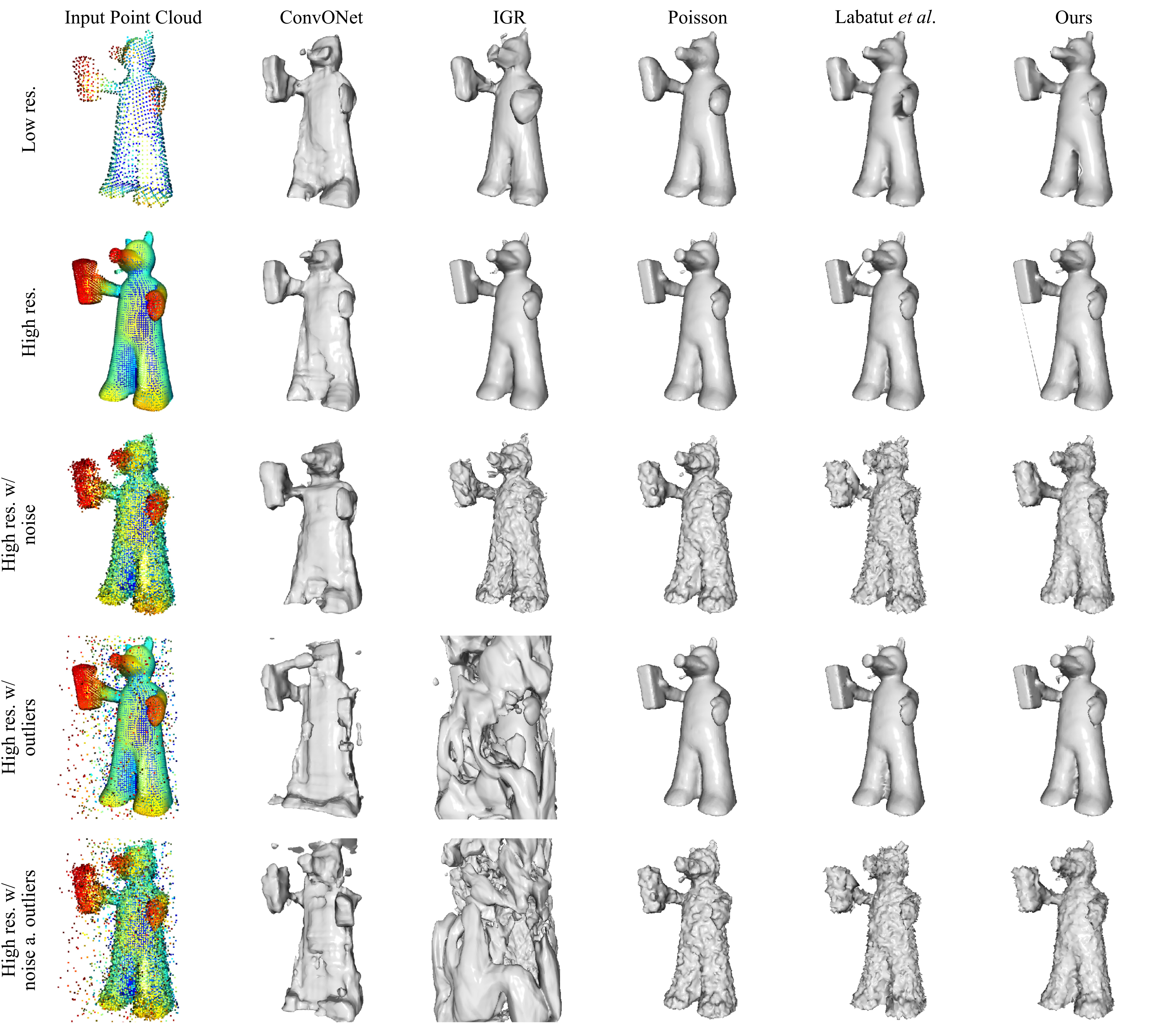}
\caption{\textbf{Reconstruction of the \emph{Quasimoto} object in the surface reconstruction benchmark of Berger \etal\ \cite{Berger2011}.} We show the input point clouds in column~1. ConvONet \cite{Peng2020} (column~2) does not generalize well to the unseen new shape. IGR \cite{Gropp2020} (column~3) is not able to filter outliers in the scan. The Screened Poisson \cite{screened_poisson} algorithm (column~4) does not reconstruct the sharp features well. The reconstructions of Labatut \etal \cite{Labatut2009a} (column~5) and ours (column~6) are visually similar for the defect-free cases. Both methods produce small artifacts in the high resolution case: between the book and nose for Labatut \etal \cite{Labatut2009a} and between the book and left foot for ours. Both methods are very robust against outliers.     
}
\label{fig:lordquas}
\end{figure*}

\noindent\textbf{Additional Qualitative Results.}
The main paper provides both quantitative results over the whole dataset (see Table~\ref{tab:berger}) and qualitative results for one object (see Fig.~\ref{fig:reconbench}). Figures \ref{fig:anchor}-\ref{fig:lordquas} show the results for all the other objects.

\begin{table*}[]
\centering
\begin{tabular}{@{}l|cccc|cccc@{}}
\toprule
\multirow{2}{*}{} & \multicolumn{4}{c|}{\multirow{2}{*}{F1-score - uncleaned mesh}} & \multicolumn{4}{c}{\multirow{2}{*}{F1-score - cleaned mesh}} \\
 & \multicolumn{4}{c|}{} & \multicolumn{4}{c}{} \\
scene & Poisson & Vu et al. & Jan. et al. & Ours & Poisson & Vu et al. & Jan. et al. & Ours \\ \midrule
kicker & 0.75 & \textbf{0.79} & 0.75 & 0.76 & 0.75 & \textbf{0.81} & 0.78 & 0.78 \\
pipes & 0.77 & \textbf{0.79} & 0.77 & 0.76 & 0.77 & \textbf{0.78} & 0.77 & 0.75 \\
delivery\_area & 0.69 & 0.70 & 0.66 & \textbf{0.71} & 0.69 & 0.70 & 0.68 & \textbf{0.71} \\
meadow & 0.45 & 0.52 & 0.51 & \textbf{0.58} & 0.40 & 0.50 & 0.50 & \textbf{0.60} \\
office & 0.60 & \textbf{0.65} & 0.59 & 0.59 & 0.60 & \textbf{0.64} & 0.62 & 0.58 \\
playground & 0.61 & \textbf{0.70} & 0.63 & \textbf{0.70} & 0.60 & 0.69 & 0.66 & \textbf{0.73} \\
terrains & 0.73 & \textbf{0.78} & 0.76 & 0.75 & 0.74 & \textbf{0.78} & 0.77 & 0.76 \\
terrace & 0.79 & 0.76 & 0.74 & \textbf{0.83} & 0.79 & 0.79 & 0.78 & \textbf{0.85} \\
relief & 0.72 & 0.67 & 0.64 & \textbf{0.80} & 0.73 & 0.69 & 0.67 & \textbf{0.80} \\
relief\_2 & 0.70 & 0.68 & 0.67 & \textbf{0.79} & 0.71 & 0.70 & 0.70 & \textbf{0.78} \\
electro & 0.65 & 0.64 & 0.60 & \textbf{0.68} & 0.65 & 0.65 & 0.64 & \textbf{0.69} \\
courtyard & 0.76 & 0.75 & 0.72 & \textbf{0.77} & 0.75 & 0.75 & 0.74 & \textbf{0.77} \\
facade & 0.50 & 0.52 & 0.50 & \textbf{0.53} & 0.51 & \textbf{0.55} & 0.54 & 0.50 \\ \midrule
mean & 0.67 & 0.69 & 0.66 & \textbf{0.71} & 0.67 & 0.69 & 0.68 & \textbf{0.71} \\ \bottomrule
\end{tabular}%
\caption{\textbf{Detailed quantitative results on ETH3D.} F1-score of all scenes of the train dataset of ETH3D \cite{Schops2017} for uncleaned and cleaned mesh reconstructions at distance $\tau = 5$\,cm. The best (highest) values per scene are in bold. We perform better than all competing methods on 8 scenes out of 13. On average, our method performs between 2 and 5\% better than the competing methods, and improve the F1-score for 8 out of $13$ scenes. The mesh cleaning only improves the F1-score of the reconstruction of Jancosek \etal \cite{Jancosek2014}.}
\label{tab:eth_f1-score}
\end{table*}


\section{Large-scale Scene Reconstruction}
\label{sec:scene}

\noindent\textbf{Metrics.}
For the large-scale benchmark ETH3D, we evaluate the mesh reconstruction methods at a given precision $\tau$ using the the Accuracy (precision) $P(\tau)$, the Completeness (recall) $R(\tau)$, and the F1-Score $F(\tau)$, defined as their harmonic mean:
\begin{align}
    F(\tau) = \frac{2 P(\tau) R(\tau)}{P(\tau) + R(\tau)}
    \label{eq:f-score}
\end{align}
We use the ETH3D Evaluation Program \cite{Schops2017} to compute these values from the ground-truth LiDAR scans and samplings of the meshed surfaces. 
In the original benchmark, the authors evaluate MVS reconstructions with threshold $\tau$ as low as $1$\,cm. Generating such mesh samplings implies sampling over $300$ million points for some scenes. To accelerate this procedure, we only sample $900$ points per $m^2$ on the reconstructed meshes. This allows us to compute accuracy and completeness with a threshold of $5$\,cm and up. 



\noindent\textbf{Detailed quantitative Results}
In \tabref{tab:eth_f1-score}, we show the 
F1-Score at $\tau = 5$\,cm of all $13$ scenes of the ETH3D dataset for both uncleaned and cleaned mesh reconstructions. 
Our method produces the best reconstruction scores for 9 out of 13 scenes. Mesh cleaning did not significantly alter the scores as it resulted in less complete but more accurate reconstructions.

\begin{figure*}[ht!]
\begin{tabular}{ccc}
     \begin{subfigure}{.32\linewidth}
       \includegraphics[width=1\linewidth]{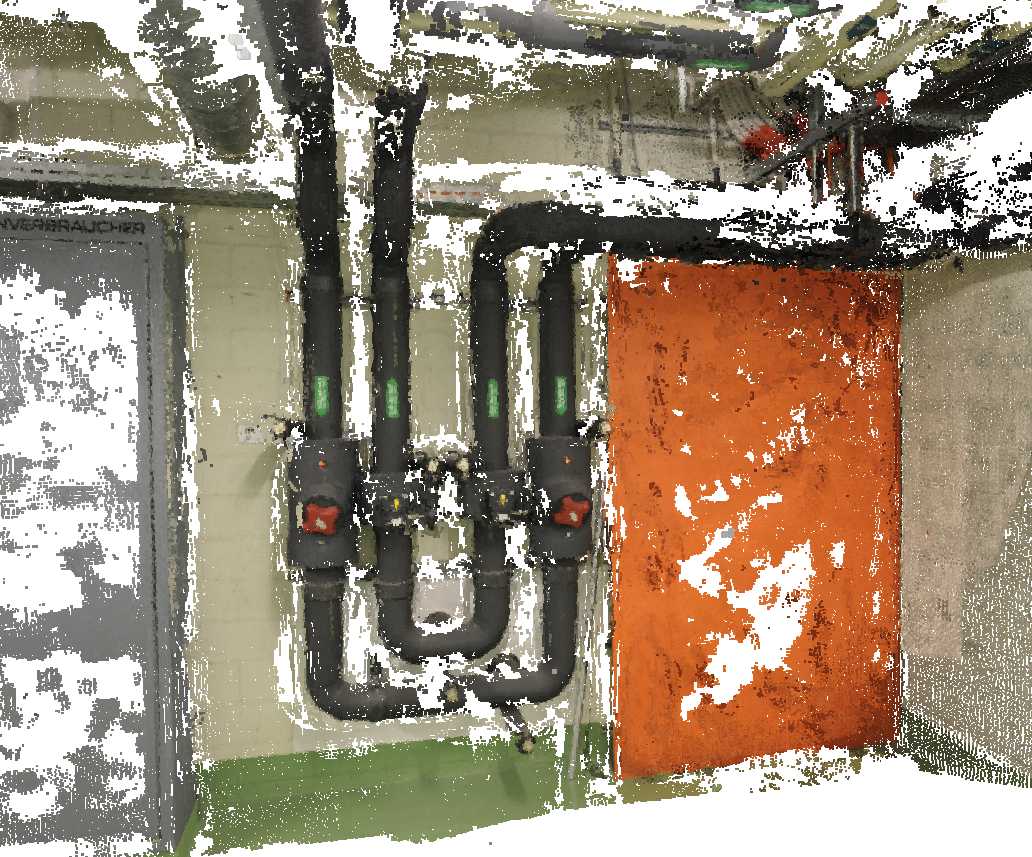}
       \caption{Dense MVS input.}
       \label{fig:occ1}
     \end{subfigure}
       \begin{subfigure}{.32\linewidth}
       \includegraphics[width=1\linewidth]{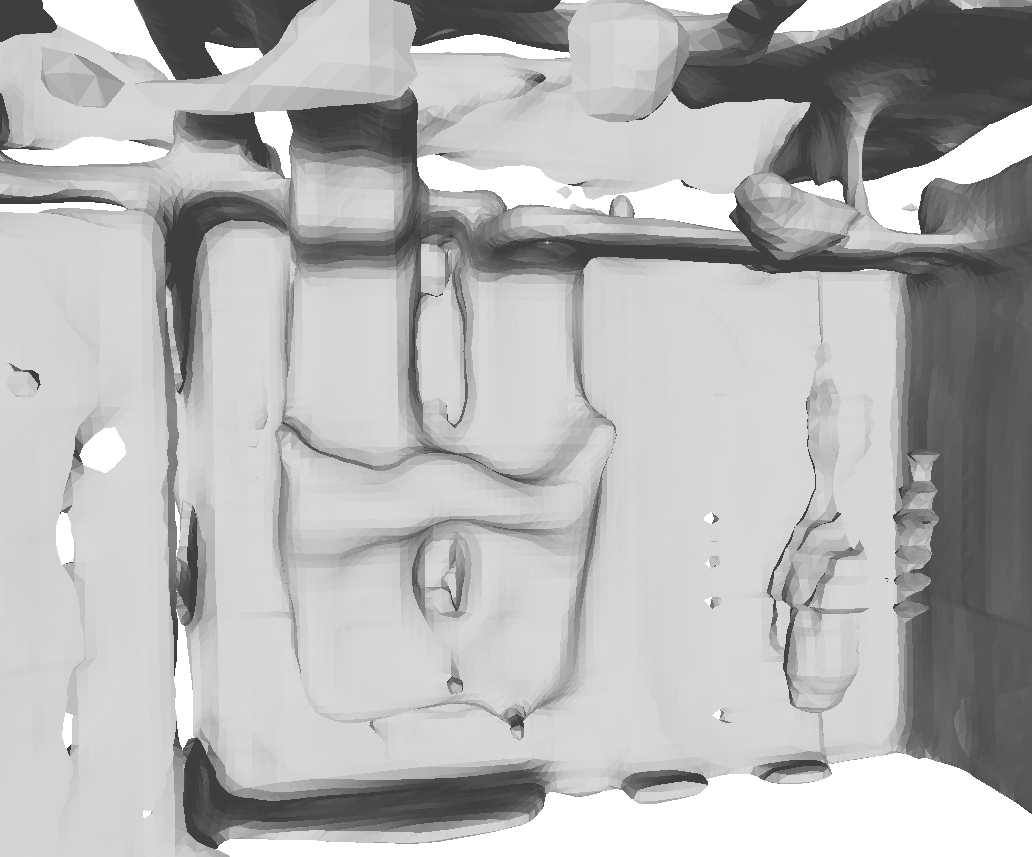}
        \caption{ConvONet.}
         \label{fig:occ2}
        \end{subfigure}
       \begin{subfigure}{.32\linewidth}
       \includegraphics[width=1\linewidth]{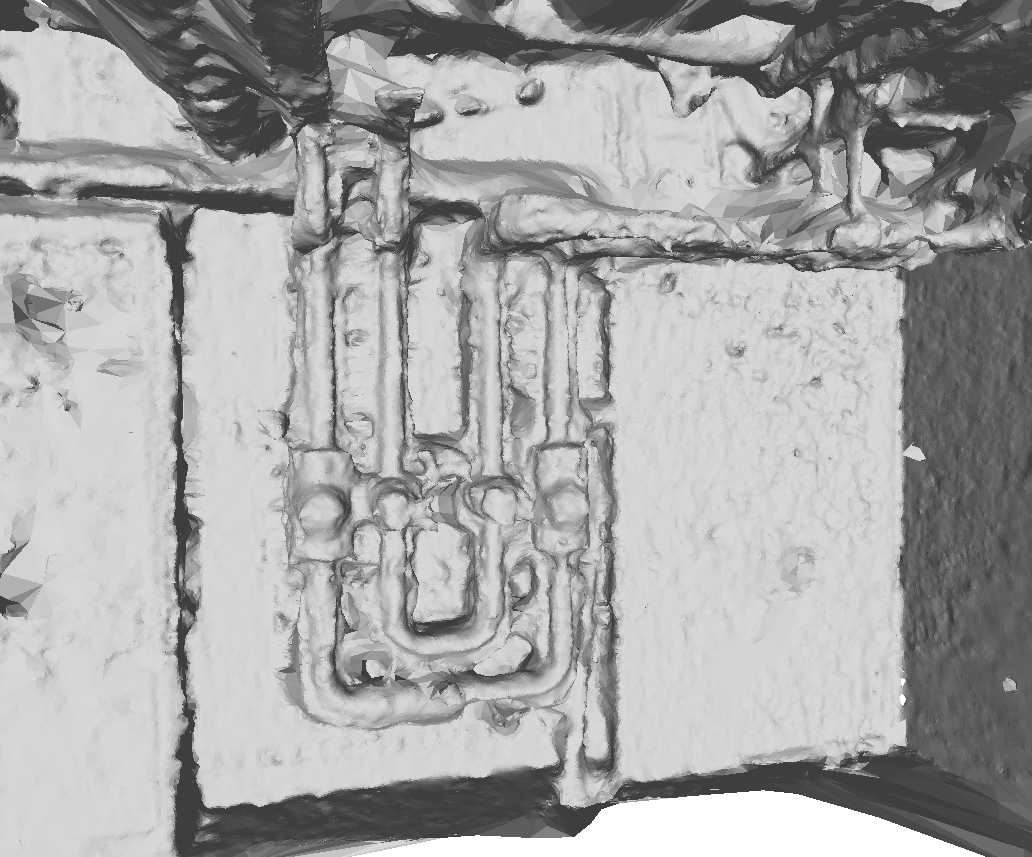}
        \caption{Ours.}
         \label{fig:occ3}
        \end{subfigure}
\end{tabular}
\caption{\RAPH{\textbf{Indoor ETH3D reconstruction.} Reconstruction of the \emph{pipes} scene of the ETH3D benchmark \cite{Schops2017}. We show the dense MVS point cloud in \Subref{fig:occ1}, the mesh reconstructions obtained by ConvONet \cite{Peng2020} in \Subref{fig:occ2} and our proposed reconstruction in \Subref{fig:occ3}. Similar to object-level reconstruction, ConvONet does not generalize well to the unseen new shapes in this scene. Our learning algorithm, operating purely locally, is able to reconstruct the pipes and fill all holes in the point cloud acquistion.}}
\label{fig:pipes}
\end{figure*}

\begin{figure*}[h]
\centering
\includegraphics[width=1\linewidth]{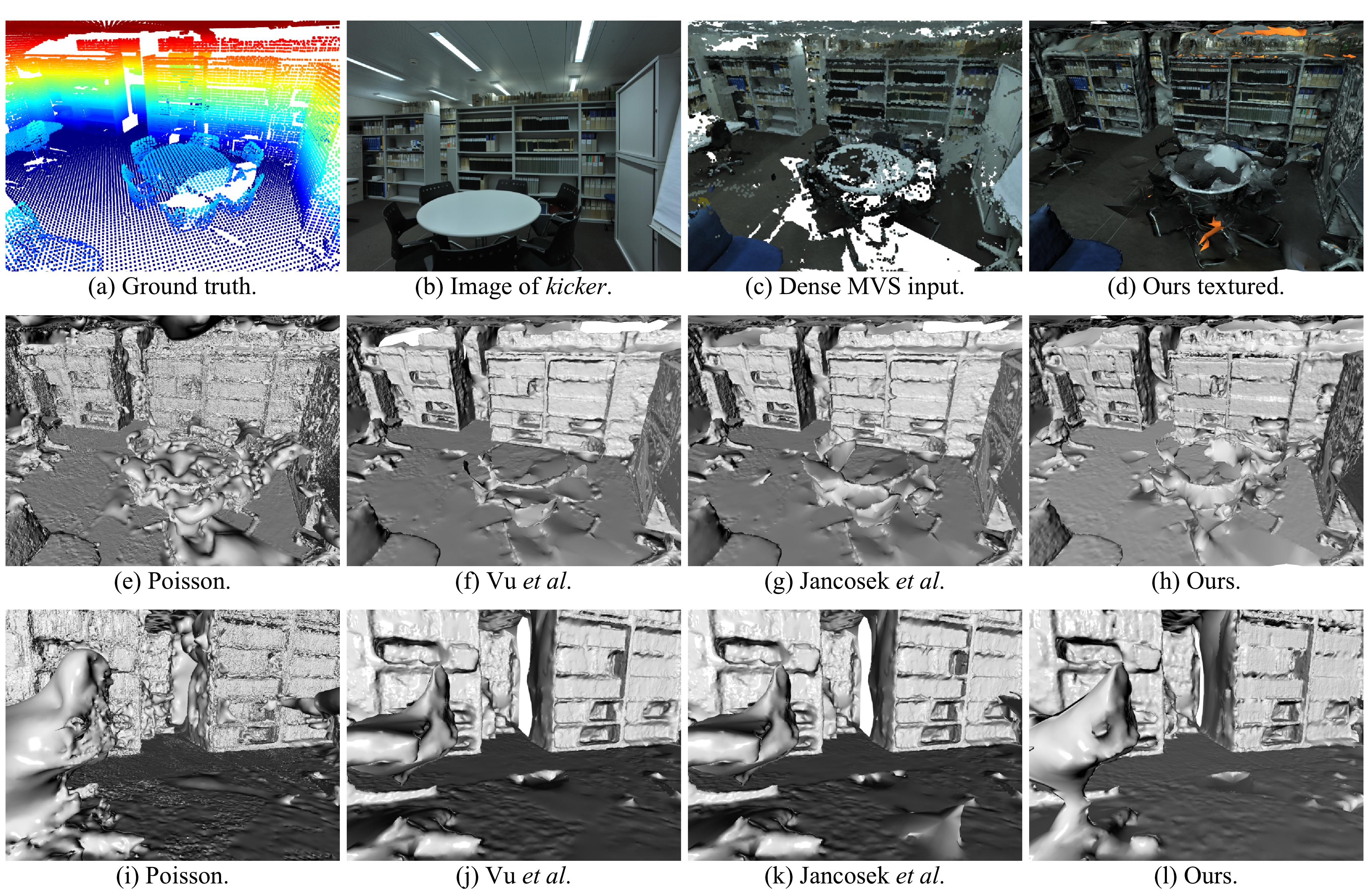}
\caption{\textbf{Indoor ETH3D reconstruction.} Reconstruction of the \emph{kicker} scene of the ETH3D benchmark \cite{Schops2017}. We show the ground truth that is used for evaluation in (a). A set of images, such as the one represented in (b), is transformed into a dense MVS point cloud (c), from which a mesh can be reconstructed and textured \cite{lettherebecolor}, as shown in (d) with our proposed mesh reconstruction. We show the untextured mesh reconstructions obtained by the screened Poisson algorithm in (e,i), the algorithms of Vu \etal \cite{Vu2012} in (f,j) and of Jancosek \etal \cite{Jancosek2014} in (g,k), and finally our proposed reconstruction in (h,l). All methods struggle to reconstruct the table and the chairs, that have little data support.
}
\label{fig:kicker}
\end{figure*}

\begin{figure*}[h]
\centering
\includegraphics[width=1\linewidth]{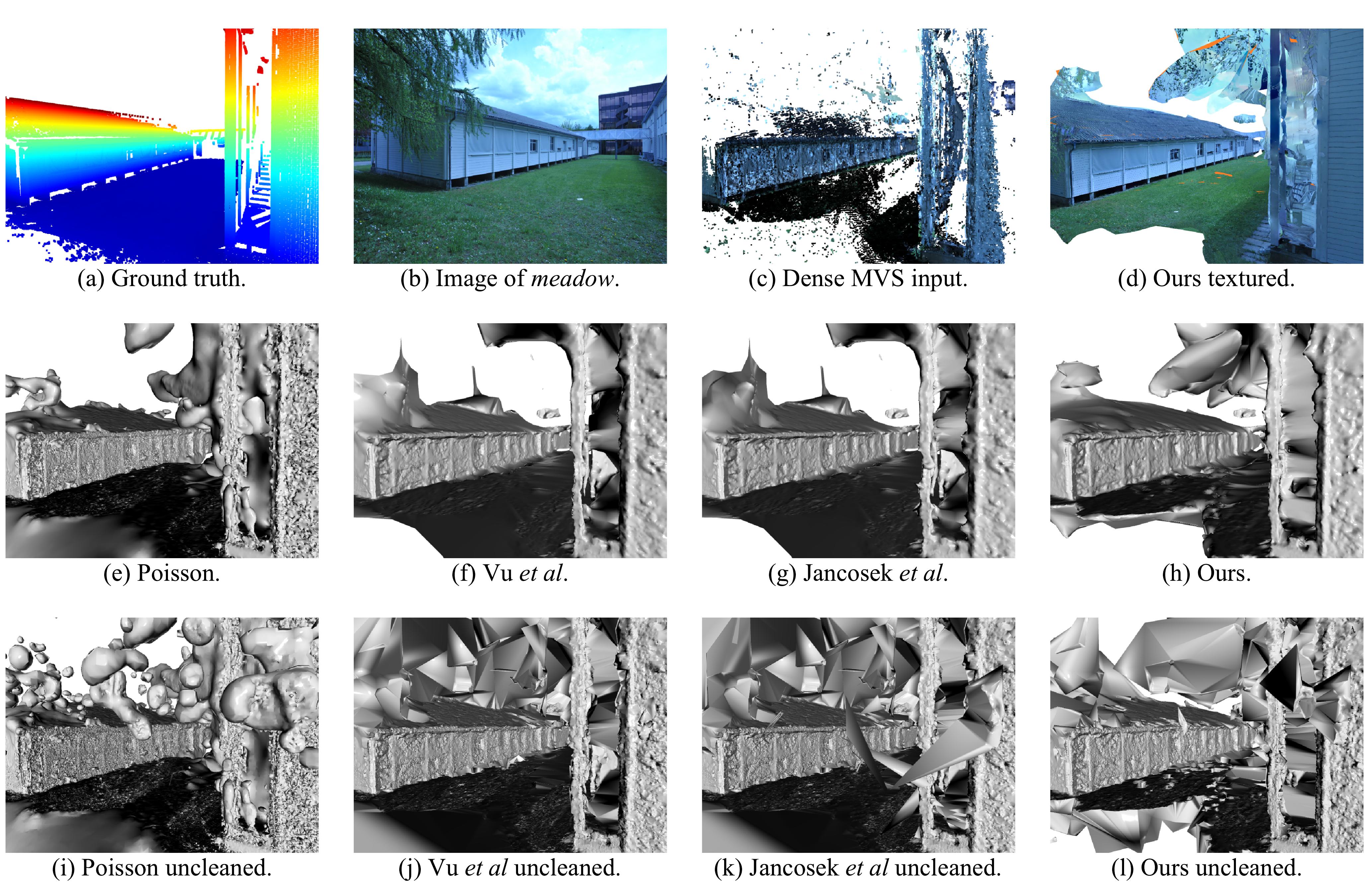}
\caption{\textbf{Outdoor ETH3D reconstruction.} Reconstruction of the \emph{meadow} scene of the ETH3D benchmark \cite{Schops2017}. We show the ground truth that is used for evaluation in (a). A set of images, such as the one represented in (b), is transformed into a dense MVS point cloud (c), from which a mesh can be reconstructed and textured \cite{lettherebecolor}, as shown in (d) with our proposed mesh reconstruction. We show the untextured mesh reconstructions obtained by the screened Poisson algorithm in (e,i), the algorithms of Vu \etal \cite{Vu2012} in (f,j) and of Jancosek \etal \cite{Jancosek2014} in (g,k), and finally our proposed reconstruction in (h,l). Trees and outliers in the sky lead to a large number of isolated components in all mesh reconstructions. Most of these small components can be removed with the heurestic mesh cleaning step that we apply as post-processing.}
\label{fig:meadow}
\end{figure*}

\begin{figure*}[h]
\centering
\includegraphics[width=1\linewidth]{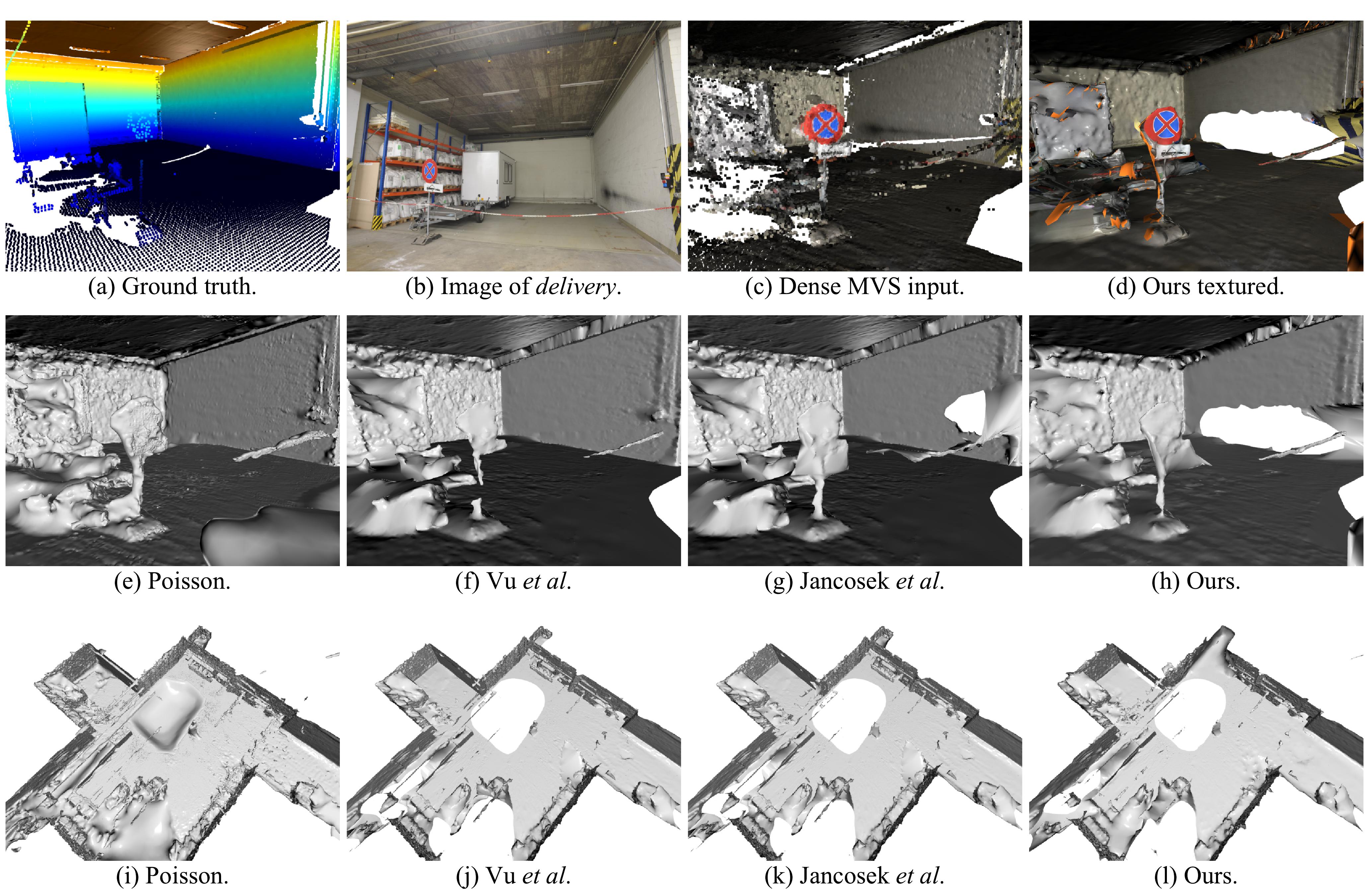}
\caption{\textbf{Failure case on ETH3D.}
Reconstruction of the \emph{delivery area} scene of the ETH3D benchmark \cite{Schops2017}. We show the ground truth that is used for evaluation in (a). A set of images, such as the one represented in (b), is transformed into a dense MVS point cloud (c), from which a mesh can be reconstructed and textured \cite{lettherebecolor}, as shown in (d) with our proposed mesh reconstruction. We show the untextured mesh reconstructions obtained by the screened Poisson algorithm in (e,i), the algorithms of Vu \etal \cite{Vu2012} in (f,j) and of Jancosek \etal \cite{Jancosek2014} in (g,k), and finally our proposed reconstruction in (h,l).
Our method does not close the wall on the right, but performs slightly better on reconstructing the no-parking sign. Yet, considering the whole scene, the holes we create do not cover a larger area than other methods.
}
\label{fig:delivery_area}
\end{figure*}

\noindent\textbf{Qualitative Results.}
%
We show an example of a locally more accurate reconstruction of our method compared to our competitors in \figref{fig:pipes} and \figref{fig:kicker}. We show in \figref{fig:meadow} the effect of the cleaning step on a hard problem due to a large amount of noise and outliers. Finally, we also show an example of our method producing a less complete reconstruction in \figref{fig:delivery_area}.

\end{document}